\documentclass[lettersize,journal]{IEEEtran}
\usepackage{amsmath,amsfonts}
\usepackage{algorithmic}
\usepackage{array}
\usepackage[caption=false,font=normalsize,labelfont=sf,textfont=sf]{subfig}
\usepackage{textcomp}
\usepackage{stfloats}
\usepackage{url}
\usepackage{verbatim}
\usepackage{graphicx}
\usepackage{cite}
\usepackage[ruled,linesnumbered]{algorithm2e}
\allowdisplaybreaks[4]
\usepackage{comment}
\usepackage{multirow}
\usepackage{makecell}
\usepackage{amsthm}
\usepackage{booktabs} 
\usepackage{tabularx}
\usepackage{amssymb}
\usepackage{balance}

\newtheorem{definition}{\bf{Definition}}
\newtheorem{theorem}{\bf{Theorem}}
\newtheorem{lemma}{\bf{Lemma}}
\newtheorem{assumption}{\bf{Assumption}}
\newtheorem{proposition}{\bf{Proposition}}

\begin{document}

\title{The Power of Bias: Optimizing Client Selection in Federated Learning with Heterogeneous Differential Privacy}

\author{Jiating Ma, Yipeng Zhou,~\IEEEmembership{Member,~IEEE,} Qi Li,~\IEEEmembership{Senior Member,~IEEE,} Quan Z. Sheng,~\IEEEmembership{Member,~IEEE,} Laizhong Cui,~\IEEEmembership{Senior Member,~IEEE,} Jiangchuan Liu,~\IEEEmembership{Fellow,~IEEE}
\thanks{Jiating Ma and Laizhong Cui are with the College of Computer Science and Software Engineering, Shenzhen University, Shenzhen 518060, China, and also with the Guangdong Laboratory of Artificial Intelligence and Digital Economy (SZ),
Shenzhen University, Shenzhen 518060, China (E-mail: 1270901086@qq.com, cuilz@szu.edu.cn).}
\thanks{Yipeng Zhou and Quan Z. Sheng are with the School of Computing, Faculty of Science and Engineering, Macquarie University, Macquarie Park, NSW 2119, Australia (E-mail: yipeng.zhou@mq.edu.au, michael.sheng@mq.edu.au).}
\thanks{Qi Li is with the Institute for Network Sciences and Cyberspace, Tsinghua University, Beijing 100190, China, and also with the Beijing National Research Center for Information Science and Technology, Beijing 100190, China (E-mail: qli01@tsinghua.edu.cn).}
\thanks{Jiangchuan Liu is with the School of Computing Science, Simon Fraser University, Burnaby, BC V5A 1S6, Canada (E-mail: jcliu@cs.sfu.ca).}
\thanks{The corresponding author is Laizhong Cui.}}

\markboth{Journal of \LaTeX\ Class Files,~Vol.~14, No.~8, August~2021}%
{Shell \MakeLowercase{\textit{et al.}}: A Sample Article Using IEEEtran.cls for IEEE Journals}


\maketitle

\begin{abstract}
   To preserve the data privacy, the federated learning (FL) paradigm emerges in which clients only expose model gradients rather than original data for conducting model training. To enhance the protection of model gradients in FL, differentially private federated learning (DPFL) is proposed which incorporates differentially private (DP)  noises to obfuscate gradients before they are exposed.  Yet, an essential but largely overlooked problem in DPFL is the heterogeneity of clients'  privacy requirement, which can vary significantly between clients and extremely complicates the client selection problem in DPFL. In other words,  both the data quality and the influence of DP noises should be taken into account when selecting clients.  To address this problem, we conduct convergence analysis of DPFL under heterogeneous privacy, a generic client selection strategy, popular DP mechanisms and convex loss. Based on convergence analysis, we formulate the client selection problem to minimize the value of loss function in DPFL with heterogeneous privacy, which is a convex optimization problem and can be solved efficiently. Accordingly, we propose the DPFL-BCS (biased client selection) algorithm. The extensive experiment results with real datasets under both convex and non-convex loss functions indicate that DPFL-BCS can remarkably improve model utility compared with the SOTA baselines. 
\end{abstract}


\section{Introduction}
\label{section:introduction}
Artificial intelligence empowered by machine learning has achieved an unprecedented success. 
However,  the advancement of machine learning gives rise to  concerns on data privacy infringement  because training complex machine learning models heavily relies on data extensively collected from clients \cite{al2019privacy}. As reported in  \cite{7093125},  user privacy can be easily inferred from collected data. To preserve data privacy during model training, the federated learning (FL) paradigm emerges, in which clients only exchange their model gradients rather than original data with a parameter server (PS) for multiple global iterations to complete model training \cite{yang2019federated}. 

However, exposing model gradients is still susceptible to malicious attacks. As investigated in \cite{jere2020taxonomy}, if model gradients are excessively exposed to attackers, FL clients may suffer from membership inference attacks \cite{DBLP:conf/sp/NasrSH19} and  reconstruction attacks \cite{yang2022using}. To defend against these attacks, differentially private federated learning (DPFL) is proposed which injects noises generated by differentially private (DP) mechanisms to distort gradients before their exposure. Although DP noises can protect clients from external attackers, the biggest challenge of DPFL lies in  significantly compromised model utility due to the disturbance of DP noises. According to \cite{10210511, 10.14778/3503585.3503592}, model accuracy can be lowered by more than 40\% by straightly injecting DP noises, making intelligent services useless. Significant efforts have been dedicated to improving DPFL model utility from various perspectives, such as adaptively allocating noises for each iteration \cite{huang2020dp}, probabilistically exposing large magnitude gradients for each iteration \cite{kerkouche2021compression,10210511}, optimizing the number of total exposure times \cite{10008087} and designing a personalized data transformation\cite{yang2023privatefl}.


It is worth noting that the privacy requirements among clients are inherently heterogeneous in all  aforementioned works, in which clients can set noise scales according to their own requirement, \emph{e.g.}, \cite{10.14778/3503585.3503592, 8423074,DBLP:journals/corr/abs-2110-15252}. Unfortunately, none of these works take this privacy heterogeneity into account when optimizing DPFL. 
According to \cite{10032626}, the privacy requirement heterogeneity is rooted from the following fact.  An altruistic and cooperative client inclines to add small noises, while a selfish and conservative client inclines to  add large noises.
Thereby, it is likely that noise scales of different clients have a giant  discrepancy. Currently, there  exists very limited  works discussing the importance of this problem but only attempting to solve it from a heuristic perspective.  Aldaghri \emph{et al.}~\cite{DBLP:journals/corr/abs-2110-15252} studied the  heterogeneous differential privacy in federated linear
regression and proposed to tune  aggregation weights of clients based on their privacy budgets. Liu \emph{et al.} \cite{10.14778/3503585.3503592} proposed to extract the top  singular subspace of the model updates uploaded by clients with more privacy budgets to project model updates from clients with small privacy budgets before the PS aggregates them. 
Due to their heuristic design, these works without theoretical guarantees are probably far away from the optimal solution.

To address the challenge, we optimize the strategy to select clients in DPFL  according to heterogeneous privacy requirements among clients with theoretical guarantees. 
Intuitively speaking, a dedicatedly designed client selection strategy should select clients with small noises more frequently and vice verse in case that DPFL performance is impaired by large noises.  Inspired by this intuition, we propose a biased client selection framework for DPFL (which is more generic than unbiased client selection widely adopted in previous works \cite{10.14778/3503585.3503592,10008087}) so that  client selection can be flexibly adjusted in accordance with  heterogeneous differential privacy among FL clients. Then, convergence analysis of DPFL is conducted to quantify the influence of  heterogeneous differential privacy and  biased client selection on model utility under standard DP mechanisms (\emph{i.e.}, Gaussian \cite{10.1145/2976749.2978318} and Laplace mechanisms \cite{wu2020value}) and convex loss. Based on convergence analysis, we formulate the problem to minimize the loss function 
with respect to the selection probability of each client. We prove that this is a convex optimization problem and further explore how to solve this problem efficiently and accurately in practical systems. 
Inspired by our analysis, the DPFL-BCS (biased client selection) algorithm is designed, which can be deployed on a semi-honest-but-curious PS \cite{280010}. 

In summary, our work contributes to optimizing FL model utility while satisfying  heterogeneous privacy requirements on clients. More specifically, our contribution is four-folds:

\begin{itemize}
\item  By proposing a generic client selection framework,  we theoretically derive the convergence rate of DPFL, in which DP noises are injected to distort gradients according to clients' heterogeneous privacy requirements. 

\item Based on our convergence analysis, we formulate a convex optimization problem 
to minimize the convergence loss (equivalent to maximizing model utility) under heterogeneous privacy requirement. 
By solving the problem, we devise the DPFL-BCS algorithm  that can optimally select participating clients. 

\item  
We explore how to implement DPFL-BCS efficiently in practical systems by exploring methods to estimate problem-related parameters (which are essential for optimizing client selection) without incurring additional privacy loss.  

\item To demonstrate the superiority of DPFL-BCS, we conduct extensive experiments with real datasets, \emph{i.e.}, Lending Club \cite{wu2020value}, MNIST \cite{10.1145/3548606.3560694}, Fashion-MNIST \cite{10210511}, FEMNIST\cite{shejwalkar2022back} and CIFAR-10 \cite{10.1145/3548606.3560557}, under both convex and non-convex loss functions. The experimental results show that DPFL-BCS can remarkably improve model utility in comparison with the SOTA baselines. In particular,  model accuracy can be incredibly improved by 30$\sim$40\% under extremely heterogeneous privacy requirement. 

\end{itemize}

The rest of the paper is organized as follows. 
The preliminary knowledge of DP and our problem definition are  introduced in Sec.~\ref{section:preliminary}.  Sec.~\ref{section:convergence} encompasses the description of the DPFL framework and the derivation of convergence. Next, the design of DPFL-BCS algorithm to optimize the client selection strategy is proposed in Sec.~\ref{section:optimize}. Sec.~\ref{section:experiment} reports our experiment results  and the relevant works are discussed in Sec.~\ref{section:relatedwork}. Finally, we conclude our work in Sec.~\ref{section:conclusion}. 

\section{Preliminary and Problem Statement}
\label{section:preliminary}

In this section, we introduce the preliminary knowledge of DP and define the problem studied by our work. 

\subsection{Preliminary of Differential Privacy}

Let $\mathcal{M}$ denotes a DP mechanism, which actually is a randomized algorithm. $\mathcal{M}$ satisfying the $(\epsilon,\delta)$-DP can be formally defined as:  
\begin{definition}
($(\epsilon,\delta)$-Differential Privacy \cite{dwork2014algorithmic}).  Assuming that $\mathcal{D}$ and $\mathcal{D}'$ are two adjacent datasets such that $\|\mathcal{D} - \mathcal{D}'\|_1\le 1 $, the randomized algorithm $\mathcal{M}$ satisfies $(\epsilon,\delta)$-DP if for any $\mathcal{D}$ and $\mathcal{D}'$, and for any $\mathcal{S} \subseteq$ Range$(\mathcal{M})$:
\begin{align}
    {Pr}\{\mathcal{M}(\mathcal{D}) \in \mathcal{S}\} \le \exp{(\epsilon)} \times {Pr}\{\mathcal{M}(\mathcal{D}') \in \mathcal{S}\}+\delta.
\end{align}
Here, Range$(\mathcal{M})$ is the output range of $\mathcal{M}$.
\label{def:1}
\end{definition}
Here, $(\epsilon, \delta)$ is  the privacy budget, depicting the amount of privacy loss of a single query. FL clients can set $(\epsilon, \delta)$ according to their own privacy requirement, resulting in heterogeneous privacy.

Our study is based on two most popular DP mechanisms: Gaussian Mechanism (GM) and Laplace Mechanism (LM). 
Let $\mathbf{w}\in \mathbb{R}^d$ denote  input parameters of a query and  ${q}(\mathbf{w}, \mathcal{D})$  denote exact query results of $d$ dimension with  input $\mathbf{w}$ on dataset $\mathcal{D}$. The Gaussian mechanism (denoted by $\mathcal{M}_{G}$) and the Laplace mechanism (denoted by $\mathcal{M}_{L}$) distort  $q(\mathbf{w}, \mathcal{D})$  according to the  following theorems. 

\begin{theorem}
\label{THE:Gaussian}
        (Gaussian Mechanism \cite{dwork2014algorithmic}). Let $\epsilon\in(0,1)$. Given the dataset $\mathcal{D}$ and query input $\mathbf{w}\in \mathbb{R}^d$ where $d$ is the dimension of input parameters $\mathbf{w}$,  the Gaussian mechanism satisfying $(\epsilon, \delta)$-DP distorts query results with $\mathcal{M}_{G}(\mathbf{w},\epsilon)={q}(\mathbf{w}, \mathcal{D})+{\mathbf{Z}_G}$.
Here $\mathbf{Z}_G$ are the Gaussian noise terms obeying  the Gaussian distribution $\mathbb{N}(0, \sigma^2\mathbb{I}_d)$. Here  $\mathbb{I}_d$ is an identity matrix with dimension $d$, 
and $\sigma$ satisfies $\sigma \ge c\Delta q_2/\epsilon$ with $c^2>2\ln(1.25/\delta)$.
$\Delta q_2=\max_{\forall \mathbf{w}, \mathcal{D}} \|{q}(\mathbf{w}, \mathcal{D}) - {q}(\mathbf{w}, \mathcal{D'})\|_2$ is the L2-sensitivity of $q(\mathbf{w}, \mathcal{D})$.
\end{theorem}

Here $\delta$ is a small number, \emph{e.g.}, $10^{-5}$, gauging the probability that $\mathcal{M}_{G}$ fails to satisfy $\epsilon$-DP. Theorem~\ref{THE:Gaussian} can guarantee $(\epsilon, \delta)$-DP for a single query. If there are $T$ queries, the noise scale should be amplified as follows. 
\begin{theorem}
\label{THE:Comp_Gaussian}
        (Composition Rule of  Gaussian Mechanism~\cite{10.1145/2976749.2978318}). Given the total number of queries $T$, for any $\epsilon<c_1T$, the Gaussian mechanism satisfies $(\epsilon,\delta)$-DP for any $\delta>0$ if the noise scale satisfies $\sigma\ge c_2\frac{\Delta q_2\sqrt{T\ln(1/\delta)}}{\epsilon}$.
\end{theorem}

Note that  $\delta=0$ for Laplace mechanism, and thus is omitted.

\begin{theorem}
\label{THE:Laplace}
        (Laplace Mechanism \cite{dwork2014algorithmic}). Given the dataset $\mathcal{D}$ and query input $\mathbf{w}\in \mathbb{R}^d$, the Laplace mechanism satisfying $\epsilon$-DP distorts  query results with $\mathcal{M}_{L}(\mathbf{w},\epsilon)={q}(\mathbf{w}, \mathcal{D})+{\mathbf{Z}_L}.$
Here ${\mathbf{Z}_L}$ are the Laplace noise terms obeying  $Pr(\mathbf{Z}_L) = \left(\frac{\epsilon}{2\Delta q_1}\right)^d\exp({-\frac{\epsilon{\|\mathbf{Z}_L}\|_1}{\Delta q_1}})$ and $\Delta q_1=\max_{\forall \mathbf{w}, \mathcal{D}} \|{q}(\mathbf{w}, \mathcal{D}) - {q}(\mathbf{w}, \mathcal{D'})\|_1$ is the L1-sensitivity of $q(\mathbf{w}, \mathcal{D})$.
\end{theorem}

Similar to the Gaussian mechanism, the Laplace mechanism amplifies the noise scale according to the following composition rule when replying queries for $T$ times.
\begin{theorem}
\label{THE:Comp_Laplace}
        (Composition Rule of  Laplace Mechanism \cite{dwork2014algorithmic}). 
        Given the total number of queries $T$, the Laplace mechanism 
        satisfies $\epsilon$-DP if the noise terms in the Laplace mechanism obey $Pr(\mathbf{Z}_L) =\left(\frac{\epsilon}{2T\Delta q_1}\right)^d\exp({-\frac{\epsilon\|{\mathbf{Z}_L}\|_1}{T\Delta q_1}})$.
\end{theorem}

\subsection{Problem Statement}

We consider a generic DPFL system with a parameter server (PS) and  $N$ clients denoted by $\mathcal{N} = \{1,\cdots,N\}$. Similar to existing DPFL works \cite{10.14778/3503585.3503592,280010}, our study is based on a semi-honest-but-curious PS, which satisfies following properties: (1) the PS selects  clients honestly according to the client selection strategy; (2) the PS honestly aggregates model gradients collected from the selected clients; (3) the PS does not modify/destroy any information for aggregation; (4) the PS tries to derive sensitive information from received model gradients and loss values of trained models.
Let $(\epsilon_n, \delta_n)$ and $T_n$ denote the privacy budget of client $n$ and the number of times client $n$ participates in FL, respectively.
Each client $n$ owns a private  local dataset denoted by $\mathcal{D}_n$ with cardinality $D_n$. 
The objective of these clients is to collaboratively  minimize the global loss function $F(\mathbf{w}) = \frac{1}{N}\sum_{n\in\mathcal{N}}F_n(\mathbf{w})$. Here, $F_n(\mathbf{w})$ is the local loss function (or objective) of client $n$ defined as $F_n(\mathbf{w})=\frac{1}{D_n}\sum_{\zeta\in\mathcal{D}_n}f_n(\mathbf{w},\zeta)$. $\mathbf{w}\in \mathbb{R}^d$ represent model parameters to be learned with dimension $d$ and $\zeta$ represents a particular data sample. 

In FL, rather than exposing $\mathcal{D}_n$, each client only returns gradients $\mathbf{g}_n^t $  to the PS for multiple global iterations via Internet communications.
However, during model communications, malicious attackers can easily eavesdrop model gradients uploaded from clients \cite{DBLP:conf/nips/ZhuLH19}, who can then put off membership inference \cite{DBLP:conf/sp/NasrSH19} or reconstruction attacks \cite{yang2022using} towards FL clients.

To defend against attackers, DPFL resorts to injecting DP noises to distort gradients before their uploading. DPFL regards global model parameters of the last global iteration, \emph{i.e.}, $\mathbf{w}^{t-1}$ in global iteration $t$, as query input and model gradients obtained by local training as query results, which should be distorted by noises. Thus, client $n$ if selected will return $\mathbf{\hat{g}}_n^t = \mathbf{g}_n^t + \mathbf{Z}_n^t$ to the PS, where $\mathbf{Z}_n^t$ represent DP noises on client $n$ in global iteration $t$. In most existing DPFL works \cite{10.14778/3503585.3503592, 10008087}, clients are selected in a pure random manner.

\begin{figure}[!t]
    \centering
        \includegraphics[width = 0.8\linewidth]{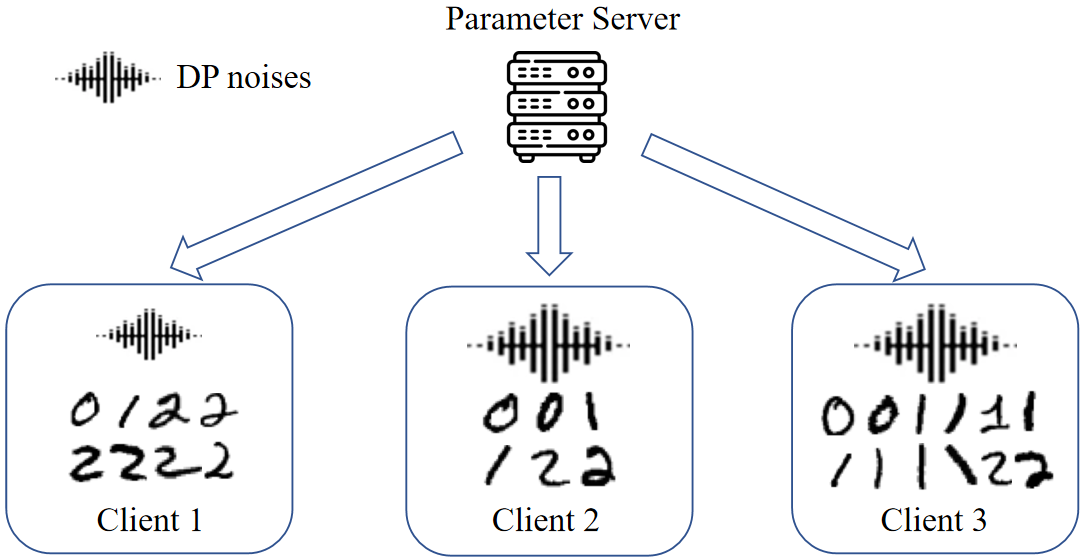}

    \caption{A specific example to illustrate the influence of heterogeneous privacy and heterogeneous data on client selection in DPFL.}
    \label{Fig:Example}
\end{figure}

\noindent{\bf An Illustrative Example.} To explain the shortcoming of these works, we show a concrete example in Fig.~\ref{Fig:Example}. 
There are three clients with heterogeneous data samples  and distinct privacy requirements. In this example, the noise scale is small on Client 1, but is much bigger on  Clients 2 and 3. From the data distribution perspective, Client 3 owns the most number of  samples while the data distribution of Client 2 is closest to the distribution of all data among all clients which is called IID. In this example, a random client selection strategy cannot discriminate values of different clients suggesting that we should at least consider: 1) The noise scale, \emph{e.g.}, the smallest noise scale on Client 1 will reduce the disturbance of noise; 2) The number of data samples, \emph{e.g.}, Client 3 owns the most number of data samples, which is vital to FL; 3) The data sample distribution, \emph{e.g.}, the IID data distribution  on Client 2  is favored for  the convergence of FL. This example indicates that client selection  is  non-trivial in DPFL with heterogeneous privacy. It is subject to both the noise scale and the data quality. The problem will be much more complicated when considering the entire FL process with multiple rounds.

\noindent{\bf Formal Problem Definition.} Suppose that $K$ clients can be selected by PS to participate in FL per global iteration. In  most existing works \cite{10008087,10.14778/3503585.3503592}, it is common to implement an unbiased strategy to randomly select clients with $T_n=\frac{KT}{N}$, where $T$ is the total number of global iterations.  This trivial strategy can guarantee the convergence of FL without DP \cite{DBLP:conf/iclr/LiHYWZ20}. However, we will theoretically prove  that  this simple selection strategy fails to attain satisfactory model utility by ignoring the impact of heterogeneous privacy. 

In our study,  we discuss a more general scenario by  only setting  $T_n\in\mathbb{N}$ and $\sum_{n\in\mathcal{N}}T_n =  KT$. 
Here, $T_n, \forall n\in \mathcal{N}$, are the most crucial variables to be optimized by our work.

 \section{Convergence Analysis with Heterogeneous Privacy}

In this section, we propose a generic DPFL framework with a biased client selection strategy so that we can analyze the convergence of DPFL under biased client selection. A summary of the key notations is presented in Appendix \ref{appendix:notation}.

\label{section:convergence}



\subsection{DPFL Framework}
\label{subsection:DPFL}

Biased client selection is not adopted in most previous DPFL works because biased selection can  lower model utility \cite{NEURIPS2023_949c57d3}. Yet,  as we have discussed in the last section, unbiased client selection may also result in inferior learning performance with heterogeneous privacy. To explore this point, we expand the client selection strategy space by including all sequences of $T_1, \dots, T_N$ satisfying $T_n\in\mathbb{N}$ and $\sum_{n\in \mathcal{N}} T_n =KT$. Let $p_n=\frac{T_n}{KT}$ denotes the selection probability of client $n$. 
Corresponding to the  huge space, we describe the DPFL framework using a generic client selection strategy as follows:
\begin{enumerate}
    \item Each client sets its privacy budget $(\epsilon_n, \delta_n)$. The PS initializes the global model $\hat{\mathbf{w}}^{0}$ and allocates the selection probability for each client $n$, \emph{i.e.}, $p_n$, subject to $\sum_{n\in\mathcal{N}} p_n=1$. Based on $p_n$, client $n$  sets the privacy budget consumed per global iteration if selected as $\epsilon_n^l = \frac{\epsilon_n}{p_nKT}$ and $\delta_n^l = \frac{\delta_n}{p_nKT}$.
    \item At the beginning of each  iteration $t\in[1,T]$, the PS selects $K$ participating clients based on $p_n$, denoted by $\mathcal{S}_t$. Then, the PS distributes the  global model $\hat{\mathbf{w}}^{t-1}$ to selected clients.
    \item Each selected  client, \emph{e.g.}, $n\in\mathcal{S}_t$,  conducts a round of local iteration with its private dataset $\mathcal{D}_n$ to update
  ${\mathbf{w}}_n^{t}\leftarrow \hat{\mathbf{w}}^{t-1} - \mathbf{g}_n^t$. Here  $\mathbf{g}_n^t = \eta_t \nabla F_n(\hat{\mathbf{w}}^{t-1}, \mathcal{D}_n)$ denote gradients and $\eta_t$ is learning rate. 
    Gradients are then distorted by $\hat{\mathbf{g}}^t_n = \mathbf{g}_n^t + \mathbf{Z}_{n}^t$, where  the noise terms $\mathbf{Z}_{n}^t$ are determined by ($\epsilon_n^l$, $\delta_n^l$) and the DP mechanisms. 
    $\hat{\mathbf{g}}^t_n$ are returned to the PS for aggregation and client $n$  updates its participation times as $C_n = C_n+1$. When $C_n\geq p_nKT$, client $n$ will not participate anymore.  
    \item The PS aggregates returned noisy gradients to  update the global model as $\hat{\mathbf{w}}^{t} \gets {\hat{\mathbf{w}}^{t-1} - \frac{1}{K}\sum_{n \in {\mathcal{S}_{t}}}\hat{\mathbf{g}}^t_n}.$ After gradient aggregation, the PS goes back to Step 2) to kick off a new round of global iteration with $\hat{\mathbf{w}}^{t}$.
\end{enumerate}

\noindent{\bf Remarks.} The biased client selection strategy can be interpreted as follows. 
 According to the aggregation rule on the PS, the expectation of aggregated gradients over $K$ selected clients is $ \mathbb{E}[\frac{1}{K}\sum_{n\in\mathcal{S}_t}\hat{\mathbf{g}}^t_n] = \frac{1}{K} * K*\sum_{n\in\mathcal{N}}p_n\hat{\mathbf{g}}^t_n=\sum_{n\in\mathcal{N}}\frac{T_n}{KT}\hat{\mathbf{g}}^t_n$. An unbiased strategy should set $T_n=\frac{KT}{N}$  to guarantee $ \mathbb{E}[\frac{1}{K}\sum_{n\in\mathcal{S}_t}\hat{\mathbf{g}}^t_n]=\frac{1}{N}\sum_{n\in\mathcal{N}}\hat{\mathbf{g}}^t_n$. This insight is correct in FL without DP, and has been widely adopted in prior works \cite{DBLP:conf/iclr/LiHYWZ20,li2020federated}. In DPFL, our study will reveal a counter-intuitive conclusion that a biased setting of $T_n$ can make model utility better if privacy is heterogeneous because $T_n$ will affect privacy consumption, and thus setting the  value of $T_n$ should take the privacy budget into account.  %

\subsection{Convergence Analysis of DPFL}
\label{subsection:convergence}
To quantify the influence of different  client selection strategies and heterogeneous noise scales on FL model utility, we  derive the convergence of DPFL  without specifying the values of $p_n$ or $T_n$. 

Before we derive the convergence rate, the influence of DP noises is quantified by their variances. 
Two most popular DP mechanisms, \emph{i.e.},  Gaussian and Laplace mechanisms, are analyzed in our work. 
Let $\mathcal{D}_n$ and $\mathcal{D}_n'$ represent two adjacent datasets. 
Gradients obtained by client $n$ in global iteration $t$ on $\mathcal{D}_n$ and $\mathcal{D}_n'$ are denoted by $\mathbf{g}_n^t=\eta_t \nabla F(\mathbf{w}_n^{t-1},\mathcal{D}_n)$ and $\mathbf{g}_n^{t'}=\eta_t \nabla F(\mathbf{w}_n^{t-1},\mathcal{D}_n')$, respectively. Then, the variances of DP noises can be represented by the following lemmas. 

\begin{lemma}\label{LEM:GauVar}(Variance of Gaussian Mechanism \cite{10.1145/2976749.2978318}) Let $\Xi_G:=\max_{\forall \mathbf{w},\forall n,\forall \zeta\in\mathcal{D}_n}\| \nabla F_n(\mathbf{w},\zeta)\|_2<\infty$ denotes the $L2$ magnitude bound of gradients and $\eta_t$ denotes the learning rate. Suppose that client $n$ is selected to participate in the $t$-th global iteration, its $L2$-sensitivity is $\Delta \mathbf{g}_{n,G} =   \frac{2\eta_t\Xi_G}{D_n}$. Let $\mathbf{Z}^t_{n,G}$ denote Gaussian noises  added to $\mathbf{g}_n^t$. The variance of $\mathbf{Z}^t_{n,G}$ is $\mathbb{E}[\| \mathbf{Z}^t_{n,G}\|^2]=\frac{4\eta_t^2\Xi_G^2dc_2^2T_n\ln(1/\delta_n)}{D_n^2\epsilon_n^2}$.
\label{lem:Gaussian_variance}
\end{lemma}

{Lemma~\ref{LEM:GauVar} can be proved based on  \cite{10.1145/2976749.2978318}. For client $n$, it is easy to obtain its $L2$-sensitivity in the $t$-th global iteration as $\| \mathbf{g}_n^t-\mathbf{g}_n^{t'} \|_2
		=\frac{\eta_t}{D_n} \| \sum_{\zeta \in \mathcal{D}_n} \nabla F_n (\mathbf{w}^{t-1},\zeta) -  \sum_{\zeta ' \in \mathcal{D}_n'} \nabla F_n (\mathbf{w}^{t-1},\zeta ')\|_2\le \frac{2\eta_t\Xi_G}{D_n}$. Then,  the variance of $\mathbf{Z}^t_{n,G}$ can be derived by substituting the sensitivity  $\Delta \mathbf{g}_{n,G} =   \frac{2\eta_t\Xi_G}{D_n}$ into  Theorem 1  in \cite{10.1145/2976749.2978318}.  }

\begin{lemma}(Variance of Laplace Mechanism \cite{10008087}) Let $\Xi_L:=\max_{\forall \mathbf{w},\forall n,\forall \zeta\in\mathcal{D}_n}\| \nabla F_n(\mathbf{w},\zeta)\|_1<\infty$ denotes the $L1$ magnitude bound of gradients and $\eta_t$ denotes the learning rate. Suppose that client $n$ is selected to participate in the $t$-th global iteration, its $L1$-sensitivity of $\mathbf{g}^t_n$ is $\Delta \mathbf{g}_{n,L}=\frac{2\eta_t\Xi_L}{D_n}$. Let   $\mathbf{Z}^t_{n,L}$ represent Laplace noises  added to $\mathbf{g}^t_n$. The variance of $\mathbf{Z}^t_{n,L}$ is $\mathbb{E}[\| \mathbf{Z}^t_{n,L}\|^2]=\frac{8d\Xi_L^2\eta_t^2T_n^2}{D_n^2\epsilon_n^2}$.
\label{lem:Laplace variance}
\end{lemma}


 To simplify our discussion, we unify the presentation of the sensitivity of $\mathbf{g}^t_n$ and  the noise terms according to Lemmas \ref{lem:Gaussian_variance} and \ref{lem:Laplace variance}. Let  $\Delta \mathbf{g}_n$ and $\mathbf{Z}^t_n$ denote the sensitivity of $\mathbf{g}^t_n$ and the  noise terms regardless of the DP mechanism, respectively. The  variance of the noise terms is denoted as  $\mathbb{E}[\| \mathbf{Z}^t_n\|^2]= \Lambda\eta_t^2T_n^z\Phi_n$, where $\Lambda=4\Xi_G^2dc_2^2$, $z=1$, $\Phi_n=\frac{\ln(1/\delta_n)}{D_n^2\epsilon_n^2}$ for GM and $\Lambda=8d\Xi_L^2$, $z=2$, $\Phi_n=\frac{1}{D_n^2\epsilon_n^2}$ for LM.

Similar to prior works \cite{cho2022towards,10008087,DBLP:conf/iclr/LiHYWZ20,NEURIPS2020_39d0a890}, we make  following assumptions to analyze the convergence of DPFL under convex loss. 

\begin{assumption}$F_n$ is L-smooth: for all $\mathbf{v}$ and $\mathbf{w}$, $F_n(\mathbf{v})\le F_n(\mathbf{w})+(\mathbf{v}-\mathbf{w})^{T}\nabla F_n(\mathbf{w})+\frac{L}{2}\|\mathbf{v}-\mathbf{w}\|^2$, $\forall n$.
\label{ass:1}
\end{assumption}
	
\begin{assumption}    $F_n$ is $\mu$-strongly convex: for all $\mathbf{v}$ and $\mathbf{w}$, $F_n(\mathbf{v})\ge F_n(\mathbf{w})+(\mathbf{v}-\mathbf{w})^{T}\nabla F_n(\mathbf{w})+\frac{\mu}{2}\|\mathbf{v}-\mathbf{w}\|^2$, $\forall n$.
    \label{ass:2}
\end{assumption} 

\begin{assumption}The variance of stochastic gradients in each client is bounded: $\mathbb{E}\| \nabla F_n(\mathbf{w},\mathcal{D}_n) -\nabla F_n(\mathbf{w}) \|^2\le \sigma^2$, $\forall n$.\label{ass:3}
\end{assumption}
Similar to \cite{cho2022towards},  to analyze the influence of selection bias, we define  selection skew to quantify the difference between a biased selection strategy and the random selection strategy. 
\begin{definition}
    (Selection skew \cite{cho2022towards}). For any model $\mathbf{w}$ and any sequence of $T_n, \forall n\in\mathcal{N}$, we define
	\begin{align}
\rho(\mathbf{w},T_1,\cdots,T_N) 
  =\frac{\sum_{n\in\mathcal{N}}\frac{T_n}{KT}[F_n(\mathbf{w})-F_n^*]}{F(\mathbf{w})-\frac{1}{N}\sum_{n\in\mathcal{N}}F^*_n}.
	\end{align} \label{def:rho}
\end{definition}
\begin{definition}(Lower bound of selection skew \cite{cho2022towards}).
    For any model $\mathbf{w}$ and any sequence of  $T_n, \forall n\in \mathcal{N}$, we define 
	\begin{align}
	\rho_{min} \triangleq \mathop{\min}_{\mathbf{w},T_1,\cdots,T_N} \rho(\mathbf{w},T_1,\cdots,T_N).
	\end{align}
\label{def:rho_min}
\end{definition}
In FL, the data distribution among clients is non-IID. The non-IID degree for each client $n$ can be quantified by the difference of loss achieved by global optimal model and local optimal model, which is  $\Gamma_n = F_n(\mathbf{w}^*)-F_n^*$. Here $\mathbf{w}^*$ and $F_n^*$ represent global optimal model and minimum local loss, respectively.
The convergence of DPFL is presented as follows.

\begin{theorem}(Convergence with Decaying Learning
	Rate). Let Assumptions \ref{ass:1}-\ref{ass:3} hold. Let learning rate $\eta_t=\frac{2}{\mu(t+\gamma)}$ and $\gamma=\frac{8L}{\mu}$.  Given a selection strategy (defined by $T_n$),
 the convergence of the DPFL framework after $T$ global iterations is:
	\begin{align}
 \label{EQ:ConvRate}
		&\mathbb{E}[F(\hat{\mathbf{w}}^T)]-F^*
		\le \frac{1}{\gamma+T}\frac{L\gamma}{2}\| \hat{\mathbf{w}}^0-\mathbf{w}^*\|^2+\frac{1}{\gamma+T}\frac{4L\sigma^2}{\mu^2K}\notag\\
  +&\underbrace{\frac{1}{(\gamma+T)T}\frac{4L^2}{K\mu^2}\sum_{n\in\mathcal{N}}T_n\Gamma_n}_{\textit{Non-IID}}
        +\underbrace{\frac{3L}{2\mu}\sum_{n\in\mathcal{N}}(\frac{T_n}{KT}-\frac{\rho_{min}}{N})\Gamma_n}_{\textit{Biased Selection}}\notag\\+&\underbrace{\frac{1}{(\gamma+T)T}\frac{4L\Lambda}{K^2\mu^2}\sum_{n\in\mathcal{N}}T_n^{z+1}\Phi_n}_{\textit{Noise Influence}},
	\end{align}where  $\Lambda=4\Xi_G^2dc_2^2$, $z=1$, $\Phi_n=\frac{\ln(1/\delta_n)}{D_n^2\epsilon_n^2}$ for GM and $\Lambda=8d\Xi_L^2$, $z=2$, $\Phi_n=\frac{1}{D_n^2\epsilon_n^2}$ for LM.
  \label{the:convergence}
\end{theorem}
Briefly speaking, Theorem~\ref{the:convergence} is proved by including  the quantified influence of the biased client selection (which can be derived based on $\rho_{min}$) and the influence of DP noises (defined in Lemmas~\ref{LEM:GauVar} and \ref{lem:Laplace variance}) into convergence rate analysis. 
The detailed proof is presented in Appendix \ref{proof_theorem}. Through the convergence analysis, we have the following insights:  
\begin{itemize}
    \item The client selection probability $\frac{T_n}{KT}$ appears in three terms: non-IID, biased selection and noise influence in the convergence rate. If the noise influence is omitted, an unbiased selection with $T_n = \frac{KT}{N}$ makes $\rho_{min}=1$ vanishing the biased selection term. This is a special scenario in our study. Due to the existence of the noise influence term, a uniform selection with $T_n = \frac{KT}{N}$ cannot guarantee a small value of the noise influence term. 
    \item If the  data distribution among clients is IID implying that $\Gamma_n = 0$, both the non-IID and biased selection terms vanish. 
    We only need to minimize  noise influence  by tuning $T_n$ to achieve the highest model utility. 
    In  distributed machine learning, it is possible to make the data distribution on workers IID \cite{NEURIPS2018_17326d10}, which can significantly simplify the client selection problem with heterogeneous privacy. However, this problem is very complicated in FL because IID data distribution cannot be achieved  \cite{DBLP:conf/iclr/LiHYWZ20} and we need to consider both the noise influence and the data distribution to maximize model utility. 
\end{itemize}

\section{Optimizing Client Selection}
\label{section:optimize}

In this section, we formulate the client selection problem in Sec.~\ref{subsection:formulation}. Then, we explore how to solve the problem in  practical systems in  Sec.~\ref{subsection:Optimized_Solution}. More discussion about our solution is represented in Sec.~\ref{subsection: Discussion}.

\subsection{Optimizing Convergence Rate}

\label{subsection:formulation}

Given an arbitrary selection strategy (defined by $T_n$), Theorem~\ref{the:convergence} proves the upper bound of  the gap between the expected global loss  and the optimal global loss   after $T$ global iterations. Thus,  
it is easy to formulate the problem to maximize the model utility of DPFL after $T$ global iterations,  equivalent to minimizing the upper bound defined in Eq.~\eqref{EQ:ConvRate}. Here $T_n$ can be regarded as variables to be optimized.

To focus on optimizing $T_n$, we ignore all terms without $T_n$ in the upper bound. Then, we formulate the problem to minimize this upper bound after $T$ global iterations. 
For simplicity, we regard all parameters not related to $T_n$ as constants by letting $\Omega_A=\frac{1}{(\gamma+T)T}\frac{4L\Lambda}{K^2\mu^2}$ and $\Omega_B = \frac{1}{(\gamma+T)T}\frac{4L^2}{K\mu^2}+\frac{1}{T}\frac{3L}{2K\mu}$. Then the objective function to optimize client selection is $\mathcal{J}(T_1, \cdots, T_N) = \Omega_A\sum_{n\in\mathcal{N}}T_n^{z+1}\Phi_n+\Omega_B\sum_{n\in\mathcal{N}}T_n\Gamma_n$, and the problem becomes:
\begin{align}
	&	\mathbb{P}_1: \min_{T_1,\cdots,T_N} \mathcal{J}(T_1, \cdots, T_N)\notag\\
		s.t.  &\sum_{n\in\mathcal{N}}T_n=KT, \quad T_n\in \mathbb{N} , \forall n\in \mathcal{N},\notag
\end{align}where $z=1$, $\Phi_n=\frac{\ln(1/\delta_n)}{D_n^2\epsilon_n^2}$ for the Gaussian mechanism and $z=2$, $\Phi_n=\frac{1}{D_n^2\epsilon_n^2}$ for the Laplace mechanism.



\subsection{Solution}
\label{subsection:Optimized_Solution}
If we relax the constraint by letting $T_n$ be a non-negative real number,  it is easy to prove that   $\mathcal{J}(\cdot)$  is a convex function.

\begin{theorem}
    $\mathcal{J}(T_1,\cdots,T_N)$ is a convex function with respect to $T_n , \forall n\in\mathcal{N}$,  under either  GM or LM.
    \label{the:convex_function}
\end{theorem}

The proof of Theorem \ref{the:convex_function} is immediate since the Hessian Matrix of function $\mathcal{J}(\cdot)$ is a semi positive-definite matrix. The detailed proof is presented in Appendix \ref{appendix: proof_of_convex_function}.


However, solving $\mathbb{P}_1$ is non-trivial in practice because we need the knowledge of problem-related parameters such as $\gamma$, $\mu$, $L$, and $\Gamma_n$  for exactly solving this problem.  In FL, the PS does not have the knowledge of these parameters prior to model training. In view of this challange,  we propose two solutions to solve $\mathbb{P}_1$: approximately optimizing $T_n$ by ignoring $\Gamma_n$ and exactly optimizing $T_n$ with estimated problem-related parameters. Both solutions are needed because the approximate solution can be derived without the knowledge of problem-related parameters, which can be used by the PS before the accurate solution is derived.
\subsubsection{Approximate Solution}


Based on our convergence analysis, if $\Gamma_n=0, \forall n\in\mathcal{N}$ (implying the data distribution is IID), it becomes easy to solve $\mathbb{P}_1$ without the necessity to estimate problem-related parameters. The sequence of  $T_n$ derived in this case is an approximate solution of $\mathbb{P}_1$, which can be denoted as follows.

\begin{proposition}
\label{Prop:ApproxSolu}
If $\Gamma_n=0, \forall n\in\mathcal{N}$, the solution to minimize $\mathcal{J}(\cdot)$ in $\mathbb{P}_1$ is $T_{n,0}^* =\frac{KT}{\sum_{n'\in\mathcal{N}}(\frac{\Phi_n}{\Phi_{n'}})^{\frac{1}{z}}}, \forall n\in\mathcal{N}$, where $z= 1$, $\Phi_n=\frac{\ln(1/\delta_n)}{D_n^2\epsilon_n^2}$ for  GM and $z=2$, $\Phi_n=\frac{1}{D_n^2\epsilon_n^2}$ for  LM.
\end{proposition}

The solution in Proposition \ref{Prop:ApproxSolu} is calculated by using Lagrange Multiplier, with details provided in Appendix \ref{appendix: proof_of_ApproxSolu}.

\begin{algorithm}[t]
\caption{Differentially Private Federated Learning with Biased Client Selection (DPFL-BCS)-PS.}
  \label{alg:Differentially Private Federated Learning-PS}
\KwIn{$T$: the total global iterations; $K$: the number of selected clients in each global iteration; $T_0$: the total global iterations for parameter estimation. }
$p_n, \forall n\in\mathcal{N}$: the selection probability of client $n$. \\
    \LinesNumbered
     \SetKwBlock{PS}{PS executes:}{}
     \PS{
     Initialize $\mathcal{A} = \mathcal{N}$, $C_n= 0, \forall n\in\mathcal{N}$, and $\hat{\mathbf{w}}^0$. \\
     \tcp{Approximate solution.}
     Initialize $p_n$ according to Proposition~\ref{Prop:ApproxSolu}.\\
    \For{$t = 1, \cdots, T$}{
      Select $K$ clients as $\mathcal{S}_t$ based on the selection probability $p_n$ from the candidate set $\mathcal{A}$.\\
     Distribute global model $\hat{\mathbf{w}}^{t-1}$.\\ 
     \For{$n \in \mathcal{S}_t$}
    {
     $\hat{\mathbf{g}}^t_n$, $\epsilon_n$ $\gets$ \textbf{Train}($n$, $t$, $\hat{\mathbf{w}}^{t-1}$, $T$, $T_0$, $p_n$, $K$).\\ 
     $C_n = C_n+1$. \tcp{Selected times.} 
     
 
     \If{$\epsilon_n\le 0$ }{
     remove $n$ from  $\mathcal{A}$. \tcp{Update $\mathcal{A}$.}}
     }
     
    Aggregation: $\hat{\mathbf{w}}^t = \hat{\mathbf{w}}^{t-1} - \frac{1}{K}\sum_{n\in \mathcal{S}_t}\hat{\mathbf{g}}^t_n$.\\
  \If{$t \leq T_0$ and $n \in \mathcal{S}_t$}{
     $\hat{F}_n(\hat{\mathbf{w}}^{t-1})$, $\hat{F}_n(\mathbf{w}^t_n)$ $\gets$ \textbf{Loss}($n$, $\hat{\mathbf{w}}^{t-1}$).
    }
     \If{$t = T_0$ }{ 
       Calculate $\Lambda$ and $\Phi_n, \forall n \in\mathcal{N}$, 
 by Eq.~(\ref{Lambda_Phi_n}). \\
 \tcp{Parameter estimation.}
 Estimate $\Gamma_n, \forall n\in\mathcal{N}$, by Eq.~(\ref{Gamma_n}).\\
 Estimate $\rho_{min}$ by Eq.~(\ref{rho_min}).\\
 Estimate $\gamma$, $L$, and $\mu$ by solving  $\mathbb{P}_2$.\\
 \tcp{Optimized solution.}
Get $T_{n}^*, \forall n\in\mathcal{N}$, by solving  $\mathbb{P}_1$.\\
    { Update} $p_n\gets \frac{T_n^*}{K(T-T_0)}, \forall n\in\mathcal{N}$.\\
    } 
    } 
    }
\end{algorithm}
\subsubsection{Optimized Solution}
We temporarily use the approximate solution in Proposition~\ref{Prop:ApproxSolu} as the client selection  strategy during the first stage. Suppose that the first stage will last $T_0$ global iterations, during which PS can estimate the problem-related parameters 
according to their definitions:
\begin{itemize}
   \item  $\Lambda$ and $\Phi_n$: In Theorem~\ref{the:convergence}, $\Lambda$ and $\Phi_n$ have been defined, which can be directly calculated by 
   \begin{align}
		 \left\{
		\begin{array}{cl}	\Lambda=4\Xi_G^2dc_2^2 \text{ and } \Phi_n=\frac{\ln(1/\delta_n)}{D_n^2\epsilon_n^2},  & \text{GM;} \\
        \Lambda=8d\Xi_L^2  \text{ and } \Phi_n=\frac{1}{D_n^2\epsilon_n^2},  & \text{LM.} 
        \end{array} \right.
        \label{Lambda_Phi_n}
\end{align}
Here $d$ is the model dimension.  $\epsilon_n$, $\delta_n$ and $D_n$ are reported by client $n$. $\Xi_G$, $\Xi_L$ and $c_2$ are the hyper-parameters. 

    \item  $\Gamma_n$: According to its definition, $\Gamma_n^t$ is estimated by  $\hat{\Gamma}_n^t = |\hat{F}_n(\hat{\mathbf{w}}^{t-1}) - \hat{F}_n(\mathbf{w}^t_n)|$ where $ \hat{F}_n(\hat{\mathbf{w}}^{t-1}) $ and $\hat{F}_n(\mathbf{w}^t_n)$ are noisy local loss values reported by client $n\in\mathcal{S}_t$ in the $t$-th iteration. Let $\mathcal{S} = \mathcal{S}_1\cup\cdots\cup\mathcal{S}_{T_0}$ denotes the union set including all clients selected in the first $T_0$ iterations. To get a reliable estimation, we estimate  $\Gamma_n$ for client $n\in\mathcal{S}$ by:
    \begin{align}
        \hat{\Gamma}_n =  \min_{1\le t\le T_0}|\hat{F}_n(\hat{\mathbf{w}}^{t-1}) - \hat{F}_n(\mathbf{w}^t_n)|
        \label{Gamma_n},  n\in\mathcal{S}.
    \end{align}
  
For other clients $n\in\mathcal{N}-\mathcal{S}$, we use  the average of the above values as the estimation. 

 \item  $\rho_{min}$: According to Definition~\ref{def:rho_min}, exactly calculating $\rho_{min}$ is complicated. We simply use  $\hat{\mathbf{w}}^{T_0-1}$ (the best model we can get until iteration $T_0-1$) to estimate $\rho_{min}$. The minimum loss $F_n^*$ should be close to $0$, and thus we let $F_n^* = 0, \forall n$. For client $n\in\mathcal{N}-\mathcal{S}_{T_0}$,  we use the average loss $\hat{F}(\hat{\mathbf{w}}^{T_0-1}) = \frac{1}{K}\sum_{n\in\mathcal{S}_{T_0}}\hat{F}_n(\hat{\mathbf{w}}^{T_0-1})$ as its local loss. Let $C_n$ be the counter to record the number of  times client $n$ is selected in the first $T_0-1$ global iterations. Then, $\rho_{min}$ is estimated  by:
 
    \begin{align}
        \hat{\rho}_{min} = \sum_{n\in\mathcal{N}}\frac{C_n\hat{F}_n(\hat{\mathbf{w}}^{T_0-1})}{K(T_0-1)\hat{F}(\hat{\mathbf{w}}^{T_0-1})}.
        \label{rho_min}
    \end{align}

 
 
    \item $\gamma$, $L$ and $\mu$:
    These parameters can be estimated  with the knowledge of $\Lambda$, $\Phi_n$, and $\hat{\Gamma}_n$ and $\hat{\rho}_{min}$.  The minimum global $F^*$ should be close to $0$, and thus we let $F^* = 0$. Let $C_n$ denote  the number of  times client $n$ is selected in the first $T_0-1$ global iterations.  After $T_0-1$ global  iterations,    Theorem~\ref{the:convergence} can bound the global loss as:   
    \begin{align}
     \label{eq:est}
		\!\!\!\!\!&\tilde{F}(\hat{\mathbf{w}}^{T_0-1})\approx\frac{\| \hat{\mathbf{w}}^0-\mathbf{w}^*\|^2}{\gamma+(T_0-1)}\frac{L\gamma}{2}-\frac{3L\hat{\rho}_{min}}{2\mu N}\sum_{n\in\mathcal{N}}\hat{\Gamma}_{n}\notag\\
  +&\frac{1}{\gamma+(T_0-1)}\frac{4L\sigma^2}{\mu^2K}
  +\frac{\sum_{n\in\mathcal{N}}C_n^{z+1}\Phi_n}{[\gamma+(T_0-1)](T_0-1)}\frac{4L\Lambda}{K^2\mu^2}\notag\\
        +&\frac{\sum_{n\in\mathcal{N}}C_n\hat{\Gamma}_n}{T_0-1}[\frac{1}{\gamma+(T_0-1)}\frac{4L^2}{K\mu^2}+\frac{3L}{2K\mu}].
	\end{align}

 In Eq.~\eqref{eq:est}, unknown parameters include $ \| \hat{\mathbf{w}}^0-\mathbf{w}^*\|^2$, $\gamma$, $\sigma^2$, $L$ and $\mu$. Meanwhile, the real global loss 
$\hat{F}(\hat{\mathbf{w}}^{T_0-1})$ can be obtained by averaging noisy local loss values $\hat{F}_n(\hat{\mathbf{w}}^{T_0-1})$ reported by the selected clients $n\in\mathcal{S}_{T_0}$. To estimate these parameters, we formulate the problem:
\begin{align}
		\!\!\mathbb{P}_2: &\min_{ \| \hat{\mathbf{w}}^0-\mathbf{w}^*\|^2,\gamma,\sigma^2, L,\mu }|\hat{F}(\hat{\mathbf{w}}^{T_0-1}) - \tilde{F}(\hat{\mathbf{w}}^{T_0-1})|\notag\\
		\!\!s.t.  &  \| \hat{\mathbf{w}}^0-\mathbf{w}^*\|^2 \ge 0,\gamma\ge 0,\sigma^2\ge 0,  L\ge 0,\mu\ge 0, L>\mu.\notag
	\end{align}
 It is easy to prove that the objective function in $\mathbb{P}_2$ is convex with respect to each individual variable. Thus, $\mathbb{P}_2$ can be efficiently solved by alternatively optimizating each variable. 
\end{itemize}




\begin{algorithm}[t]
\caption{Differentially Private Federated Learning with Biased Client Selection (DPFL-BCS)-Client.}
  \label{alg:Differentially Private Federated Learning-Client}
    \LinesNumbered
\SetKwBlock{Training}{Train ($n$, $t$, $\hat{\mathbf{w}}^{t-1}$, $T$, $T_0$, $p_n$, $K$):}{}
     \Training{
     Initialize $\eta_t$ in each iteration. \\
     Calculate sensitivity: $\Delta \mathbf{g}_n'$  if $t\le T_0$; else $\Delta \mathbf{g}_n$.\\
     Local training: $\mathbf{g}^t_n \gets \eta_t \nabla F_n(\hat{\mathbf{w}}^{t-1}, \mathcal{D}_n)$.\\
    \text{Calculate privacy budget consumed by each iteration: }
     When $t = 1$, $\epsilon_n^l = \frac{\epsilon_n}{KTp_n}$, $\delta_n^l = \frac{\delta_n}{KTp_n}$; when $t = T_0+1$, $\epsilon_n^l = \frac{\epsilon_n}{K(T-T_0)p_n}$, $\delta_n^l = \frac{\delta_n}{K(T-T_0)p_n}$.\\
   Generate noises $\mathbf{Z}^t_n$ with budget ($\epsilon_n^l$, $\delta_n^l$).\\
           Distort gradients: $\hat{\mathbf{g}}^t_n \gets \mathbf{g}^t_n + \mathbf{Z}^t_n$.\\
            Update the remaining budget: $\epsilon_n = \epsilon_n - \epsilon_n^l$, $\delta_n = \delta_n - \delta_n^l$. \\
           \textbf{return}  $\hat{\mathbf{g}}^t_n$, $\epsilon_n$.\\
             }
\SetKwBlock{getloss}{Loss ($n$, $\hat{\mathbf{w}}^{t-1}$): }{}
\getloss{
             Get local loss values $F_n(\hat{\mathbf{w}}^{t-1})$, $F_n(\mathbf{w}^t_n)$ with $\hat{\mathbf{w}}^{t-1}$ and $\mathbf{w}^t_n$, respectively. \\
             Generate noises $z^t_{n,1}$ , $z^t_{n,2}$ with   budget ($\epsilon_n^l$, $\delta_n^l$).\\
             \text{Distort loss values: }$\hat{F}_n(\mathbf{w}^t_n) \gets \frac{\eta_t{F}_n(\mathbf{w}^t_n)+z^t_{n,1}}{\eta_t}$,
$\hat{F}_n(\hat{\mathbf{w}}^{t-1})\gets \frac{\eta_t{F}_n(\hat{\mathbf{w}}^{t-1})+z^t_{n,2}}{\eta_t}$.
             \\\textbf{return}  $\hat{F}_n(\hat{\mathbf{w}}^{t-1})$, $\hat{F}_n(\mathbf{w}^t_n)$.
}
\end{algorithm}

With the estimation of problem-related parameters, it is easy to solve $\mathbb{P}_1$ to get the optimal $T_{n}^*$ for the second stage. Note that the remaining total number of  global iterations is $T - T_0$. Therefore, the optimal selection probability for client $n$ is $p_{n}^* = \frac{T_{n}^*}{K(T-T_0)}$.
Based on the two-stage design, the entire algorithm framework of DPFL-BCS for PS and clients are presented in Alg.~\ref{alg:Differentially Private Federated Learning-PS} and Alg.~\ref{alg:Differentially Private Federated Learning-Client}, respectively.

\subsection{Discussion}

\label{subsection: Discussion}
Computing $p_n$ to optimize client selection is not cost free. We further  discuss the cost associated with this process. 

\noindent{\bf Privacy Budget Consumption for Computing $p_n$}.  
Considering the privacy leakage risks for reporting local loss values for parameter estimation, selected clients generate DP noises to distort local loss value as well. We regard loss values, \emph{i.e.}, $\eta_t{F}_n(\hat{\mathbf{w}}^{t-1})$ and   $\eta_t{F}_n(\mathbf{w}^t_n)$, 
as two additional parameters in  model $\mathbf{w}$ for disturbance. In other words, the model dimension of $\mathbf{w}$ is expanded from $d$ to $d+2$ in the first stage.

Because we involve two additional parameters, the sensitivity function should be altered accordingly. Let $ \Theta$ denotes the maximum value of the loss function evaluated by  a single sample. 
 Then, the additional sensitivity incurred by protecting each local loss value $F_n(\mathbf{w})$ for client $n$ can be represented as   $\Delta F_n = \frac{\Theta}{D_n}$. Therefore, 
in  the $t$-th global iteration during the  first stage, selected client $n\in\mathcal{S}_t$ generates DP noises  to distort both gradients $\mathbf{g}^t_n$ and loss values, \emph{i.e.}, $\eta_tF_n(\hat{\mathbf{w}}^{t-1})$ and  $\eta_tF_n(\mathbf{w}^t_n)$, based on the  revised sensitivity:
\begin{align}
    \Delta \mathbf{g}_n' =  \left\{
		\begin{array}{cl}	\frac{2\eta_t\Xi_G+2\eta_t\Theta}{D_n},  & \text{GM;} \\
        \frac{2\eta_t\Xi_L+2\eta_t\Theta}{D_n} ,  & \text{LM.} 
        \end{array} \right.
        \label{sensitivity_loss_gradient}
\end{align}

The proof is straightforward by using the sensitivity  definition. When considering exposing two additional loss values, the total sensitivity  is bounded by $ \Delta \mathbf{g}_{n}' = \Delta \mathbf{g}_n + 2\eta_t\Delta F_n$. Then, the selected client $n$ uploads the noisy gradients $\hat{\mathbf{g}}^t_n$ and noisy loss values, \emph{i.e.}, $\hat{F}_n(\hat{\mathbf{w}}^{t-1})$ and  $\hat{F}_n(\mathbf{w}^t_n)$, to the PS.

\noindent{\bf Computation and Memory Overhead.} All problem-related parameters can be estimated after $T_0 $ global iterations. Since both $\mathbb{P}_1$ and $\mathbb{P}_2$ are convex optimization problems, it only consumes little computational resource. 
Besides, $T_0$ is a small constant as we only need $T_0$ samples to reasonably estimate these parameters.  It implies that the memory overhead is $O(N)$ on the PS side because the PS only spends $O(1)$ memory overhead for each client.

\noindent{\bf Incomplete Information with Privacy Budget}. In our study, we suppose that each client honestly reports its  privacy budget to the PS, which is  the same as scenarios studied by most related works such as  \cite{10.14778/3503585.3503592}.  
Nonetheless, it is possible that clients refuse to report their privacy budget information in practice \cite{DBLP:journals/jsac/SunCWWWWS21}. 
Based on recent studies, we briefly discuss how  DPFL-BCS can overcome this deficiency.

 In DPFL-BCS, it is possible to  add a  module to infer the privacy budgets of clients on the PS. The PS can obtain the distribution of privacy budgets among clients by making a survey questionnaire~\cite{DBLP:journals/jsac/SunCWWWWS21}. Furthermore, the PS can infer the privacy budgets of  clients from their historical training information~\cite{10032626}.

\begin{figure*}[t]
    \centering
    \includegraphics[width=\linewidth]{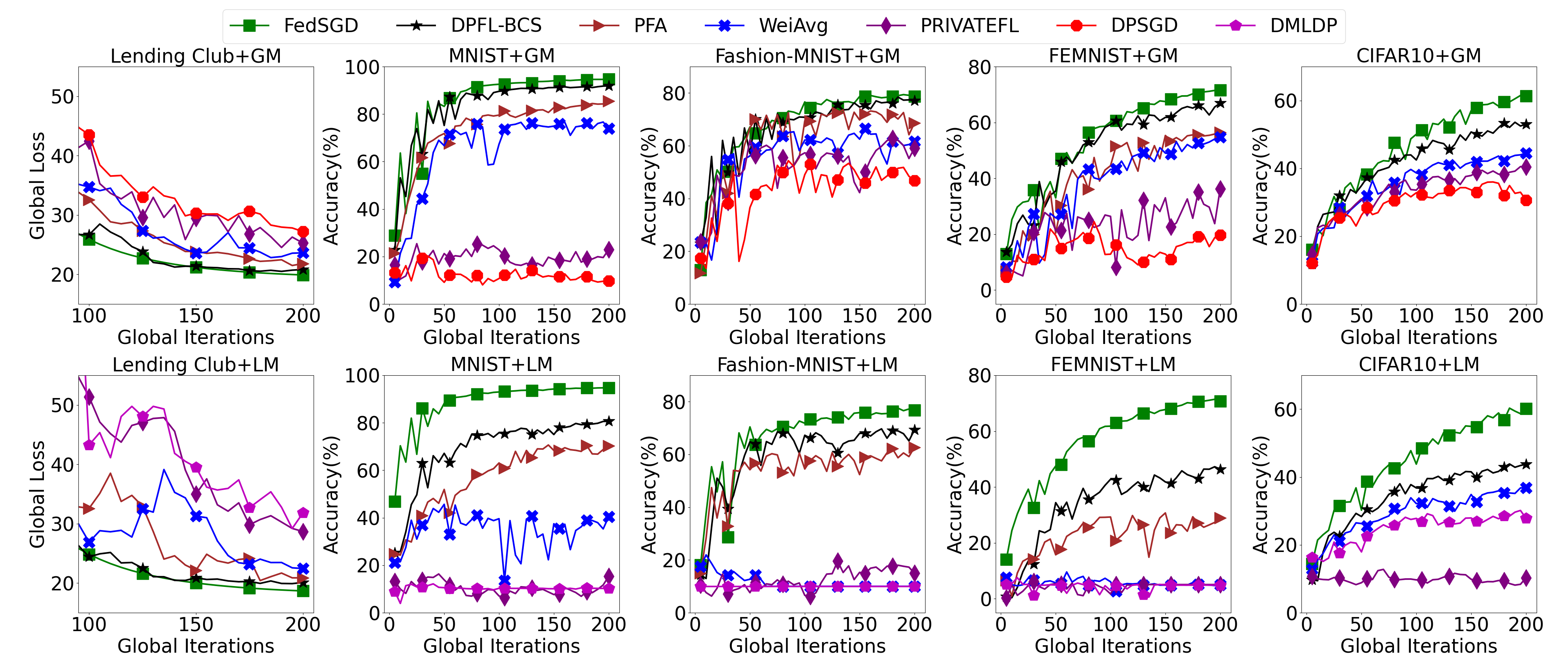}
    \caption{Model utility comparison of different algorithms under fixed privacy heterogeneity (GM = Gaussian Mechanism, LM = Laplace Mechanism).}
    \label{fig:diff_algorithm}
\end{figure*}

 \section{Performance Evaluation}
\label{section:experiment}
In this section, we report our experimental results conducted with real datasets and compared with the SOTA baselines. 

\subsection{Experimental Settings}

\noindent{\bf Datasets.} We conduct experiments  on five popular public datasets,  \emph{i.e.}, Lending Club \cite{wu2020value}, MNIST \cite{10.1145/3548606.3560694}, Fashion-MNIST \cite{10210511}, FEMNIST\cite{shejwalkar2022back} and CIFAR-10 \cite{10.1145/3548606.3560557}. 

Lending Club contains nearly 890,000 loan records collected from a peer-to-peer lending platform, which is available on Kaggle. We randomly select 40,000 records for our experiments, with 80\% as the training set and 20\% as the test set. The further processing of the Lending Club dataset is the same as that in \cite{wu2020value}. We set the interest rates of  loans per annum as the label of each record while others are used as  features. We remove  unique identifier features, such as IDs and member IDs. Features with the value of ``NULL” are removed as well before we encode categorical features. After that, we employ Principal Component Analysis (PCA) to further select  top ten important features as our training features.
Finally, we regularize each selected feature, which is conducive to the stability of the training model. 

Both MNIST  and Fashion-MNIST  contain 70,000 28*28 grayscale images which can be classified into 10 classes.  The labels of  MNIST images are digits from  0-9, while the labels of  Fashion-MNIST images are  daily goods. The CIFAR-10 dataset consists of 60,000 32*32 color images, which can also be classified into 10 categories. For each dataset, we randomly select 10,000 samples as test set and  the rest as training set. 

FEMNIST  contains 805,263 28*28 grayscale images collected from 3,500 users, which can be classified into 62 classes (10 for digits, 26 for lowercase letters and 26 for uppercase letters). We randomly sample 10\% samples from the whole dataset for conducting our experiments. Then, we randomly select 90\% samples as training set, and use the rest as test set. 

The training set will be further distributed to FL clients as private local datasets. Whereas, the test set is retained by the PS for evaluating  model performance. 

\noindent{\bf Trained Models.} We implement a  linear regression model containing 11 parameters same as the one in \cite{wu2020value} to predict the interest rate of the Lending Club records.  The loss function is defined by Mean Square Error (MSE) for Lending Club while by Cross-Entropy for other four datasets. We implement a convolutional neural network (CNN) model with two convolution layers and a fully connected layer \cite{10210511} to classify MNIST images. The model consists of 4,266 parameters. The LeNet-5 network \cite{10210511} is implemented to classify Fashion-MNIST images which contains 61,706 parameters.

We implement a CNN model similar to the one in \cite{shejwalkar2022back} with 67,026 parameters to classify FEMNIST images, which contains two convolution layers and three fully connected layers. Each convolution layer is followed by a BatchNorm2d layer, a ReLU function and a  max pooling layer.
 The ResNet-20 \cite{10.1145/3548606.3560557} is implemented  to classify CIFAR-10 images, which contains 272,474 parameters.

\noindent{\bf System Settings.} We simulate a FL system with a single PS and $N = 100$ clients. We set the total number of global iterations as $T = 200$, while the parameter estimation phase lasts $T_0 = 10$  iterations. In each iteration, the PS selects $K = 20$ participating clients to conduct local training. For all five models, each client adopts the Stochastic Gradient Descent (SGD) for local training with the momentum $0.9$ and  the weight decay rate $2\times10^{-4}$ \cite{10.1145/3548606.3560694}. We set   learning rates as $\frac{0.02}{1+\frac{t}{200}}$, $\frac{0.05}{1+\frac{t}{200}}$, $\frac{0.05}{1+\frac{t}{200}}$, $\frac{0.05}{1+\frac{t}{200}}$ and $\frac{0.2}{1+\frac{t}{100}}$ for Lending Club, MNIST, Fashion-MNIST, FEMNIST and CIFAR-10 \cite{10008087}.

We implement Gaussian and Laplace mechanisms to generate noises to obfuscate model gradients and local loss values in our experiments.  According to prior works \cite{10.1145/2976749.2978318, 10210511}, the gradients should be clipped to bound sensitivity before adding DP noises. 
We use the norm clipping for gradients to obtain the bound $\Xi_G$ and $\Xi_L$ for Gaussian  and Laplace mechanisms, respectively. 
Let $B$ denotes the clipping bound.  Each gradient is clipped by $\nabla F_n(\hat{\mathbf{w}}^{t-1}, \zeta)[j]=\frac{\nabla F_n(\hat{\mathbf{w}}^{t-1}, \zeta)[j]}{\max(1,\|\nabla F_n(\hat{\mathbf{w}}^{t-1}, \zeta)\|_2/B)}$  and $\nabla F_n(\hat{\mathbf{w}}^{t-1}, \zeta)[j]=\frac{\nabla F_n(\hat{\mathbf{w}}^{t-1}, \zeta)[j]}{\max(1,\|\nabla F_n(\hat{\mathbf{w}}^{t-1}, \zeta)\|_1/B)}$ for  Gaussian  and Laplace mechanisms. Here, $\zeta\in\mathcal{D}_n$ is a data sample and   $\nabla F_n(\hat{\mathbf{w}}^{t-1}, \zeta)[j]$ is the $j$-th item of the vector $\nabla F_n(\hat{\mathbf{w}}^{t-1}, \zeta)$. Through trials of different values, we finally set $B = 150, 20, 25, 25$ and $5$ for the Lending Club, MNIST, Fashion-MNIST, FEMNIST and CIFAR-10.  Besides, we also need to set a proper value for the maximum value of the loss function, \emph{i.e.}, $\Theta$, according to our design. This is a limited number in most machine learning problems. For example, for a classification problem with 10 labels, we can use a random guess strategy to estimate the upper bound of loss values. If the loss  is defined by the Cross-Entropy function, the loss of the random guess strategy (which is regarded as the worst strategy achieving the highest loss value) is $ \Theta = \ln{10}$. 
For the linear regression model
 predicting interest rate with Lending Club, we cap  $\Theta = 150$ according to our experiment results.

To conduct experiments under non-IID data distribution, a well-known property of FL, we allocate training data samples to clients according to  the Dirichlet distribution \cite{DBLP:conf/nips/Yu0K0J22}, in which  a parameter $\alpha$  can control the non-IID degree of data distribution on clients. A smaller $\alpha$ implies a more significant non-IID degree and vice verse.

The privacy requirements are heterogeneous, and each client uses a uniform and random method to set its privacy budget $\epsilon_n$ and $\delta_n$ from the ranges $[\epsilon_{min},\epsilon_{max}]$ and $[\delta_{min},\delta_{max}]$, respectively. Note that $\delta_{min} =\delta_{max}=0$ for the Laplace mechanism.  
By default,  we set $\delta_{min} = 10^{-5}$ and $\delta_{max} = 10^{-4}$ for the Gaussian mechanism. We set $\epsilon_{min} = 0.1$ for Lending Club and MNIST,  $\epsilon_{min} = 0.5$ for Fashion-MNIST and FEMNIST, and $\epsilon_{min} = 1$ for CIFAR-10. We vary the degree of privacy heterogeneity  by changing $\epsilon_{max}$.

\noindent{\bf Baselines.} We implement three kinds of baselines: 
\begin{itemize}
    \item FedSGD \cite{pmlr-v54-mcmahan17a} does not involve any DP mechanism to protect exposed gradients. It provides the upper bound of model utility to measure the influence of DP noises on FL performance.
    \item DPSGD~\cite{10.1145/2976749.2978318} and  DMLDP~\cite{wu2020value} adopt  Gaussian   and Laplace mechanisms to protect gradients, respectively. However, they fail to consider privacy heterogeneity among different clients, and select all clients with the same probability. 
    \item  WeiAvg  was proposed in  \cite{10.14778/3503585.3503592} to handle  privacy heterogeneity by heuristically adjusting the aggregation weight of each client based on its privacy budget. PFA is  upgraded from WeiAvg, relying on extracted top singular subspace of model gradients. Its computation overhead is prohibitive, and cannot run at all for  ResNet-20  with  272,474 parameters. 

    \item  PRIVATEFL~\cite{yang2023privatefl} was designed to locally implement a personalized data transformation  to mitigate data heterogeneity among clients in DPFL. However, it uniformly selects participating clients  because it ignores   privacy heterogeneity among clients.

\end{itemize}


To fairly compare these baselines, we conduct two groups of experiments to compare them based on Gaussian and Laplace mechanisms, respectively. Note that DPFL-BCS is a generic framework that can incorporate both Gaussian and Laplace mechanisms. 
For a given DP mechanism, we fix the privacy budget consumed by each algorithm to compare their model utility in terms of MSE  for Lending Club and model classification accuracy for other datasets.

\subsection{Experimental Results}

\subsubsection{Comparison of Model Utility under Fixed Privacy Heterogeneity}

We compare  model utility of different algorithms under fixed privacy heterogeneity. More specifically, we set  $\alpha = 3$ and $\epsilon_{max} = 3$ for Lending Club and MNIST, $\alpha = 3$ and $\epsilon_{max} = 4$ for Fashion-MNIST and FEMNIST, and $\alpha = 3$ and $\epsilon_{max} = 6$ for CIFAR-10 to allocate the local dataset and sample  the privacy budget for each client. The privacy budget is then fixed when comparing different algorithms. The training  results are plotted in Fig.~\ref{fig:diff_algorithm} with upper subfigures showing  results with  Gaussian mechanism and lower subfigures showing results with  Laplace mechanism. In each figure, the x-axis represents the number of conducted global iterations while the y-axis represents the model utility evaluated by the test set retained by the PS. From these results, we can observe that: 1) DPFL-BCS achieves the highest model utility compared with other DP based FL algorithms. Its performance is the one closest to FedSGD which does not involve DP noises to protect gradients; 2)  Compared to  Gaussian mechanism,  Laplace mechanism compromises model utility more significantly, which however can provide a stronger privacy protection; 3) Although our theoretical analysis is based on convex loss, experimental results indicate that DPFL-BCS is effective under both convex and non-convex loss functions;  4) PFA and WeiAvg outperform DPSGD and DMLDP because they have considered privacy heterogeneity in their design. Yet, their performance is still much worse than ours because they are heuristic based; 5) PRIVATEFL outperforms DPSGD with Gaussian noises because it can reduce data heterogeneity among clients to improve the utility. However, PRIVATEFL fails to improve the performance of DMLDP with Laplace noises, and even achieves worse performance on CIFAR-10. Besides, its performance is always worse than ours in each experimental case because it does not take privacy heterogeneity into account.   

 \begin{figure}
 \includegraphics[width=\linewidth]{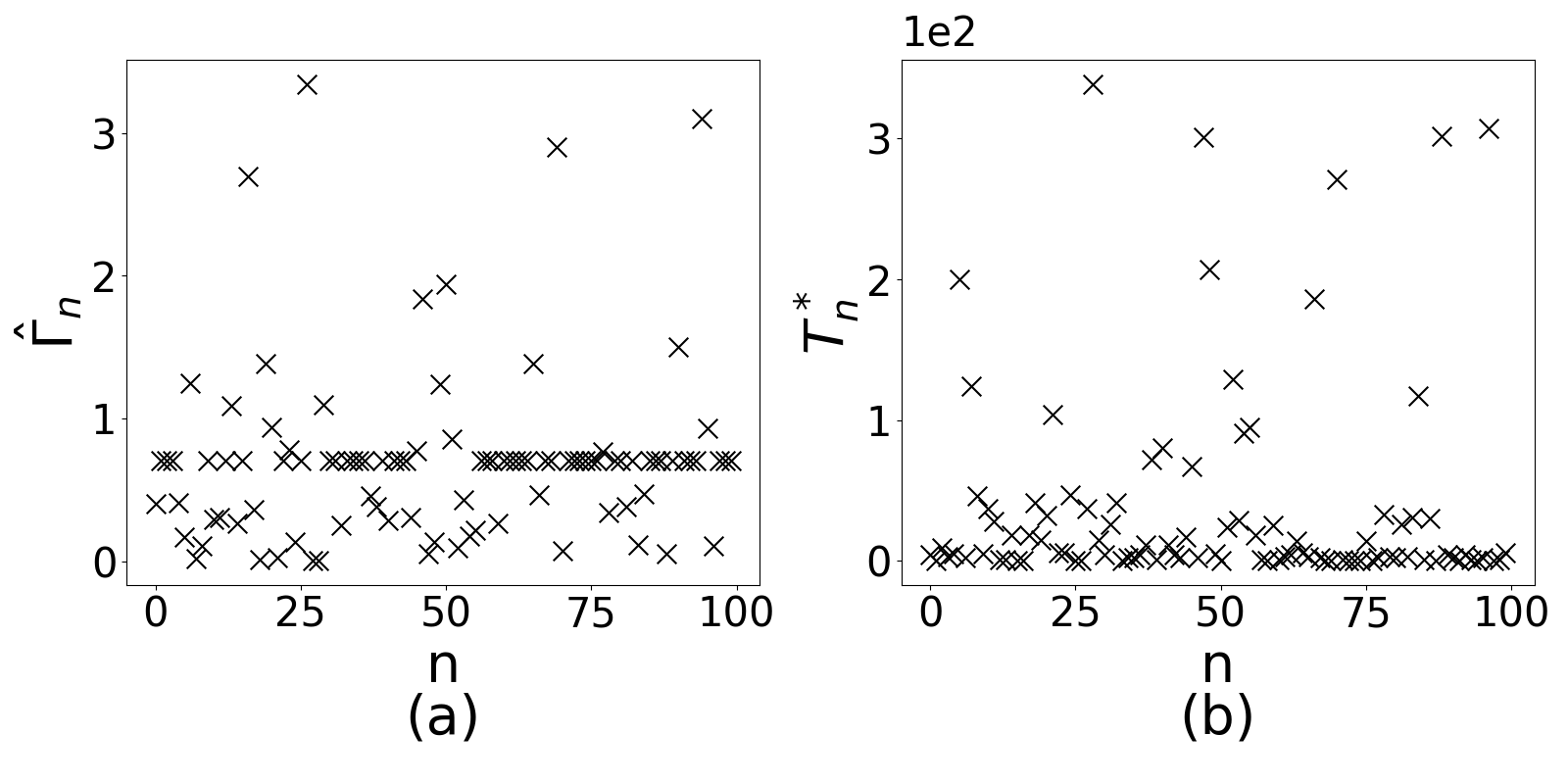}

    \caption{The estimation of problem-related parameter $\Gamma_n$ and the optimal client selection decision $T_n^*$ in MNIST with Gaussian Mechanism.}
\label{fig:mnist+Gaussian+all+est}

\end{figure}

\begin{table*}[!t]

 	\caption{Final model utility by combining DPFL-BCS and PRIVATEFL. }
 	\centering
 	\label{PrivateFL}
 	\begin{tabularx}{0.9\linewidth}{c|c|ccccc}
 		\toprule[1pt]
 		DP&Alg.&Lending Club&MNIST&Fashion-MNIST&FEMNIST&CIFAR-10\\
 		\midrule
\multirow{2}{*}{Gaussian Mechanism}&PRIVATEFL\cite{yang2023privatefl}&25.28$\pm$1.25&(22.95$\pm$0.77)\%&(57.12$\pm$2.39)\%&(49.67$\pm$1.14)\%&(40.49$\pm$0.83)\%\\

\multirow{2}{*}{}&PRIVATEFL-BCS&\textbf{20.45}$\mathbf{\pm}$\textbf{0.56}&\textbf{(91.84}$\mathbf{\pm}$\textbf{0.53)\%}&\textbf{(80.73}$\mathbf{\pm}$\textbf{1.43)\%}&\textbf{(67.05}$\mathbf{\pm}$\textbf{0.61)\%}&\textbf{(56.13}$\mathbf{\pm}$\textbf{0.52)\%}\\
\hline
\multirow{2}{*}{Laplace Mechanism}&PRIVATEFL\cite{yang2023privatefl}&28.68$\pm$1.78&(12.91$\pm$0.54)\%&(14.94$\pm$0.68)\%&(5.08$\pm$0.47)\%&(10.28$\pm$0.33)\%\\

\multirow{2}{*}{}&PRIVATEFL-BCS&\textbf{20.94}$\mathbf{\pm}$\textbf{0.63}&\textbf{(81.25}$\mathbf{\pm}$\textbf{1.29)\%}&\textbf{(69.42}$\mathbf{\pm}$\textbf{1.24)\%}&\textbf{(47.35}$\mathbf{\pm}$\textbf{1.29)\%}&\textbf{(46.41}$\mathbf{\pm}$\textbf{0.60)\%}\\
 		\bottomrule[1pt]
 	\end{tabularx}
 
 \end{table*}


\subsubsection{Visualizing Selection Decisions for Heterogeneous Privacy}

To better understand the  biased  decisions made by DPFL-BCS, we visualize a particular experiment case 
conducted with MNIST plus the Gaussian mechanism while the results of other nine cases are shown in Appendix~\ref{appendix:problem-related} due to limited space. 
Here, we only visualize the estimation of ${\Gamma}_n$ in Fig.~\ref{fig:mnist+Gaussian+all+est}(a) after the first stage, which can be used to  explain how DPFL-BCS makes biased client selection decisions. 
With the estimation of $\Gamma_n$, we can solve  problem $\mathbb{P}_2$ to obtain the estimation  of $\gamma$, $L$ and $\mu$, which are shown in Appendix~\ref{appendix:problem-related}.

Based on estimated problem-related parameters, we finally optimize $T_n$ by solving $\mathbb{P}_1$ to obtain the optimal client selection decision, \emph{i.e.}, $T_n^*, \forall n \in \mathcal{N}$.  As we observe from 
Fig.~\ref{fig:mnist+Gaussian+all+est}(b), the optimal client selection decision is indeed biased to accommodate heterogeneous  privacy budget distribution and data distribution. The results of other experiment cases are  very similar with details presented in Appendix~\ref{appendix:problem-related}.

\subsubsection{Combination with DP-Improving Algorithms} We demonstrate that it is possible to combine DPFL-BCS with other DP-improving algorithms, which have ignored clients' heterogeneous privacy requirement. 
We implement PRIVATEFL~\cite{yang2023privatefl} as a case study under  both Gaussian and Laplace mechanisms. It means that a biased client selection strategy in lieu of a pure random strategy is adopted in PRIVATEFL.  The results in Table~\ref{PrivateFL} are presented with mean $\pm$ std, which show that: 1) PRIVATEFL has poor performance under heterogeneous privacy settings because it does not discriminate  clients with heterogeneous privacy when making selection decisions; 2) DPFL-BCS can significantly improve the performance of PRIVATEFL under heterogeneous privacy by selecting clients in a biased manner. 

\subsubsection{Comparison of Model Utility by Varying Privacy Heterogeneity}

\begin{figure*}[t]
    \centering
    \includegraphics[width=\linewidth]{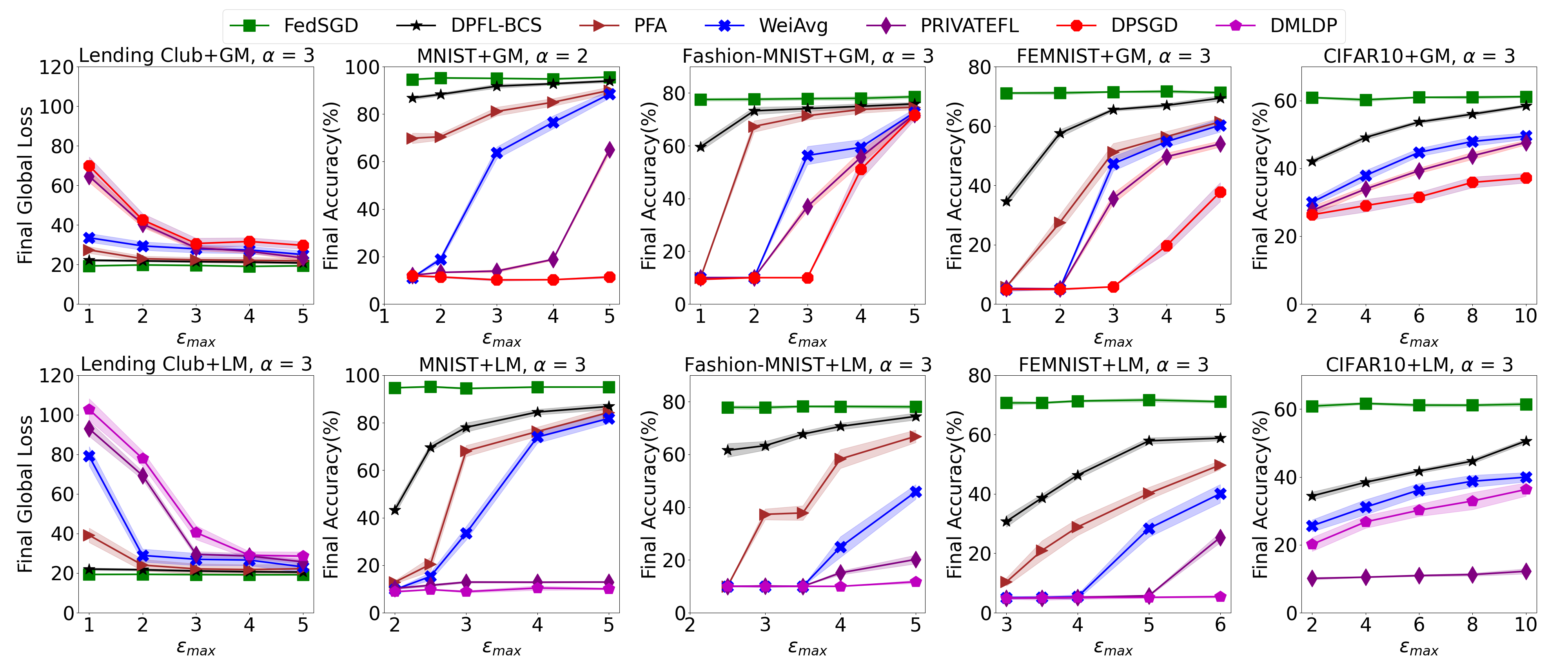}
    \caption{Final model utility comparison of different algorithms by varying $\epsilon_{max}$ (GM = Gaussian Mechanism, LM = Laplace Mechanism).}
    \label{fig:diff_e_all_alg}
\end{figure*}

To validate that DPFL-BCS can adapt with the degree of privacy heterogeneity, we vary $\epsilon_{max}$ from $2$ to $10$ for CIFAR-10, while from $1$ to $6$ for other four datasets to examine how the degree of privacy heterogeneity influences  model utility. The results presented by mean $\pm$ std are plotted in Fig.~\ref{fig:diff_e_all_alg} with the x-axis representing the privacy heterogeneity degree and the y-axis representing the final model utility. 


From the experimental results, we can observe that: 1)  Under all experiment cases with different levels of privacy heterogeneity, DPFL-BCS is always the best one which can achieve the highest model utility compared with other DP based baselines; 2) As the increase of the privacy budget upper bound $\epsilon_{max}$, model utility of DPFL-BCS, WeiAvg and PFA  gets better as these algorithms can exploit  privacy heterogeneity to improve FL performance. However, PRIVATEFL, DPSGD and DMLDP algorithms cannot effectively improve FL performance with the increase of $\epsilon_{max}$ because the privacy heterogeneity issue is ignored in their design.

\subsubsection{Comparison of Model Utility by Varying Data Heterogeneity}

\begin{figure*}[t]
    \centering
    \includegraphics[width=\linewidth]{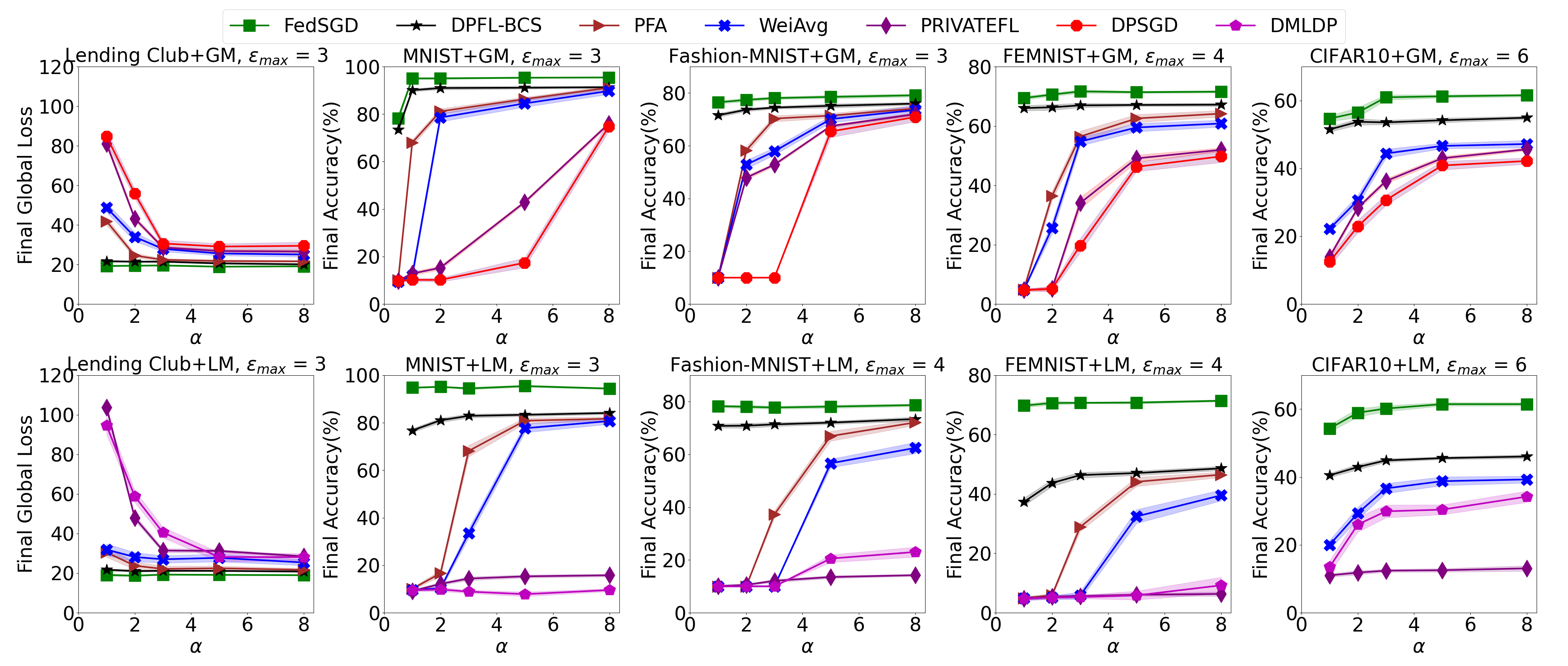}
    \caption{Final model utility comparison of different algorithms by varying $\alpha$ (GM = Gaussian Mechanism, LM = Laplace Mechanism).}
    \label{fig:diff_alpha_all_alg}

\end{figure*}

To further explore how the heterogeneity degree of data distribution influences the model utility, we conduct experiments by varying the parameter  $\alpha$ of the Dirichlet  distribution  from $0.5$ to $8$.  Then, we compare the final model utility after $T=200$ iterations between DPFL-BCS and other baselines. The experimental results with mean $\pm$ std  are presented in Fig.~\ref{fig:diff_alpha_all_alg} with the x-axis representing the non-IID degree and the y-axis representing the final model utility. 

As we can see from these results: 1) In Dirichlet  distribution, a smaller $\alpha$ implies  more significant non-IID. Thus,  the model utility deteriorates with the decrease of  $\alpha$;  2) DPFL-BCS is always the best one achieving the highest model utility among all DP baselines. The reason is that DPFL-BCS can properly balance data and privacy heterogeneity to significantly improve model utility. In particular,  it is worth noting that DPFL-BCS can notably improve 30$\sim$40\% model accuracy when $\alpha$ is extremely small; 3) WeiAvg and PFA can only work well when $\alpha$ is not very small because their heuristic design fails to optimize model utility in extreme cases.

\section{Related Work}
\label{section:relatedwork}
In this section, we  briefly overview the development of FL and its security threats. Then, we highlight our contributions in light of  recent advances in  DPFL.
\subsection{Federated Learning and Security Threats}
The FL framework was originally proposed to protect data privacy when training machine learning models \cite{yang2019federated}.  \cite{pmlr-v54-mcmahan17a} is the first one devising model average algorithms, \emph{i.e.}, FedAvg and FedSGD, for FL,  which allows decentralized clients to  only exchange model parameters (or gradients) with the PS to complete model training. The data privacy is preserved since clients never expose their raw data  to the PS. Later on, \cite{DBLP:conf/iclr/LiHYWZ20}  analyzed the convergence of FedAvg/FedSGD under non-IID data distribution. To accelerate the convergence of FL, different variants of FedAvg/FedSGD have been proposed by \cite{DBLP:conf/iclr/LiSBS20,NEURIPS2020_39d0a890} with theoretical guarantees.

Owing  to its privacy protection capability, FL has been widely used in   privacy sensitive applications, such as digital health \cite{rieke2020future} and smart cities \cite{zheng2022applications}. 
Nevertheless, as reported in  \cite{jere2020taxonomy}, FL is still susceptible to two kinds of malicious attacks: model performance attacks and data privacy attacks. The objective of model performance attacks is to demolish FL training to yield a final model with poor performance. 
For example, a data poisoning attack was proposed in  \cite{shejwalkar2022back} that can attack FL by flipping data labels on clients. Then, a more advanced attack algorithm was designed by using gradient ascent to fine-tune the global model and increase its loss on unseen data.
Our study focuses on the second kind of attacks which attempt to infer data privacy from leaked model information.  
In  \cite{DBLP:conf/sp/NasrSH19},  a white-box inference attack was developed that processes extracted gradient features from different layers of
the target model separately, and combines their information to compute the membership probability of a target data point. Zhu \emph{et al.} \cite{DBLP:conf/nips/ZhuLH19} proposed the DLG  (Deep Leakage from Gradients)  algorithm, which can reconstruct original  samples with high accuracy using leaked gradients.

\subsection{Differentially Private Federated Learning}

In order to defend against data privacy attacks targeting for model gradients in machine learning (ML), DP mechanisms were introduced, which distort private information with zero-mean random noises. Wu \emph{et al.} \cite{wu2020value} incorporated  Laplace mechanism into distributed ML  to protect the exposed gradients. 
Nonetheless, its model utility was severely compromised by DP noises since noises are  straightly injected.
Abadi \emph{et al.} \cite{10.1145/2976749.2978318} refined  Gaussian mechanism used in ML by more precisely estimating the privacy loss to reduce the scale of random noises. 
However, these works were not designed for FL, and thus did not consider non-IID data distribution.  

Later on, the adoption of DP mechanisms has been extended for  DPFL, and  efforts were made to alleviate the noise influence on model utility. Huang \emph{et al.} \cite{huang2020dp} proposed a  novel DPFL framework to adaptively allocate noises injected into gradients. Hu \emph{et al.} \cite{hu2021concentrated} applied the zCDP mechanisms for DPFL to add fewer noises compared to DP mechanisms. Considering that the noise scale is proportional to the exposed model size, compressive sensing \cite{kerkouche2021compression} and privately gradient selection \cite{10210511} techniques were introduced into DPFL by reducing the number of exposed parameters. Zhou \emph{et al.} \cite{10008087} proposed to optimize the numbers of model exposure times  to reduce the influence of DP noises. The work \cite{pmlr-v151-noble22a} incorporated DP into the popular SCAFFOLD algorithm, which is a variant of FedSGD, by taking data heterogeneity into account. Furthermore, PRIVATEFL\cite{yang2023privatefl} was proposed to reduce the client heterogeneity introduced by both data and DP noises through a carefully designed personalized transformation.

However, aforementioned works did not consider privacy heterogeneity among DPFL clients. In reality, DPFL clients set privacy budgets based their own privacy requirement, and thereby the discrepancy of noise scales between different clients is probably  significant.  
Aldaghri \emph{et al.}\cite{DBLP:journals/corr/abs-2110-15252} proposed a heterogeneous setup for privacy and theoretically analyzed the privacy-utility trade-off in federated linear regression.  Then, Liu \emph{et al.} \cite{10.14778/3503585.3503592}
proposed the WeiAvg algorithm that uses the ratio of privacy as aggregate weights to mitigate the influence from clients with large noises, but this heuristic design failed to optimize model utility.
Based on WeiAvg, the PFA algorithm was proposed which divides selected clients into two sets: ``public” and ``private” in each global iteration based on their privacy budgets. Then, it extracts the top singular subspace of model updates submitted by ``public” clients to  project  model updates of ``private” clients before aggregating them with the ratio of privacy. The computational complexity of PFA is prohibitive in practice, proportional to model dimension and client population. Its memory overhead is also significant proportional to the square of model dimension.  
In contrast, our design is based on convergence analysis, in which the client selection problem  under heterogeneous privacy is formulated into a convex optimization problem. Thus, our design is lightweight and guarantees optimized model utility in comparison to PFA. 

\section{Conclusion}
\label{section:conclusion}
Federated learning   is the most popular technique for data privacy preservation in model training. Differentially private has been widely incorporated into FL to enhance the protection level. Nevertheless, the heterogeneous privacy requirement on FL clients will compromise their data values to different extents, which has not been well addressed by existing works. To fill in this gap, we analyze the convergence of FL with DP noises to quantify the influence of DP noises on model utility. Based on convergence analysis, we optimize the client selection strategy that can properly incorporate each client into FL based on its noise scale so that the final model utility is maximized. Moreover, our design is lightweight incurring negligible overhead. Comprehensive experiments are conducted to demonstrate that our design can significantly improve model utility compared with the state-of-the-art baselines. Although our initial work proves the  effectiveness of DPFL-BCS,  
in the future work, we will explore how to extend our design under more complicated scenarios such as online FL and decentralized FL systems. 





\bibliographystyle{IEEEtran}
\bibliography{ref.bib}

\begin{IEEEbiography}[{\includegraphics[width=1in,height=1.25in,clip,keepaspectratio]{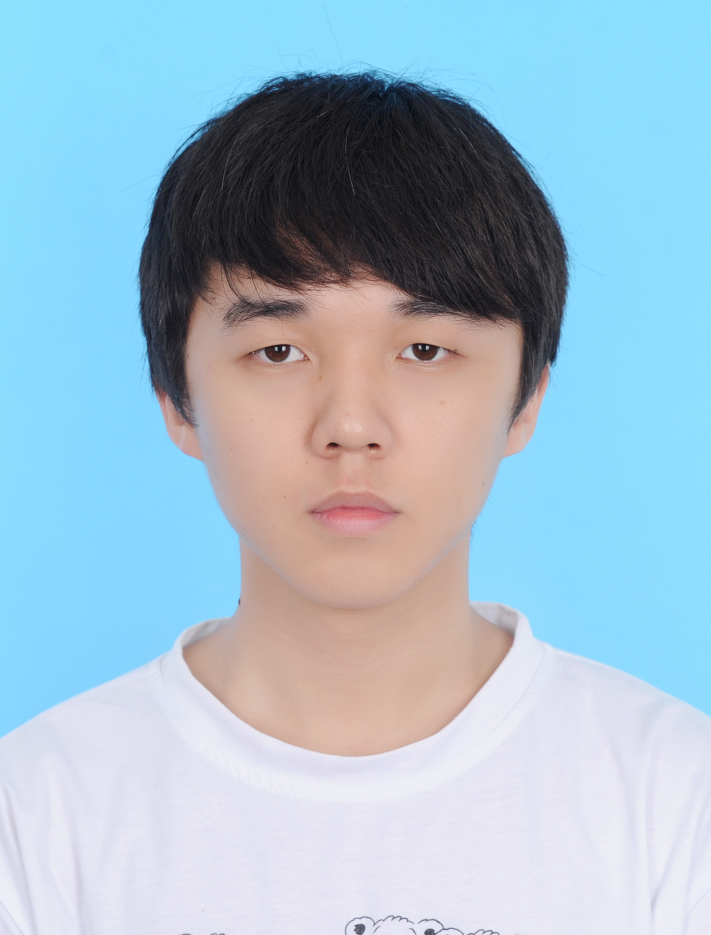}}]{Jiating Ma}  received his B.Sc. degree from Shenzhen University, Shenzhen, China, in 2021. He is pursuing his PhD degree in Shenzhen University. His research interests include federated learning and differential privacy. He has published papers in IEEE TII and IEEE TDSC.
\end{IEEEbiography}

\begin{IEEEbiography}[{\includegraphics[width=1in,height=1.25in,clip,keepaspectratio]{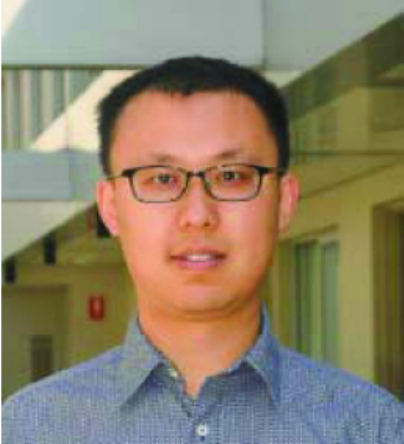}}] {Yipeng Zhou}  is a senior lecturer in computer science with School of Computing at Macquarie University, and the recipient of ARC DECRA in 2018. From Aug. 2016 to Feb. 2018, he was a research fellow with Institute for Telecommunications Research (ITR) of University of South Australia. From 2013.9-2016.9, He was a lecturer with College of Computer Science and Software Engineering, Shenzhen University. He was a Postdoctoral Fellow with Institute of Network Coding (INC) of The Chinese University of Hong Kong (CUHK) from Aug. 2012 to Aug. 2013. He won his PhD degree and Mphil degree from Information Engineering (IE) Department of CUHK respectively. He got Bachelor degree in Computer Science from University of Science and Technology of China (USTC). His research interests lie in distributed/federated learning, privacy protection and caching algorithm design in networks. He has published more than 90 papers including IEEE INFOCOM, ICNP, IWQoS, IEEE ToN, JSAC, TPDS, TMC, TMM, etc.
\end{IEEEbiography}

\begin{IEEEbiography}[{\includegraphics[width=1in,height=1.25in,clip,keepaspectratio]{Qi_Li.pdf}}]{Qi Li} received the PhD degree from Tsinghua University. He has ever worked with ETH Zurich and the University of Texas at San Antonio. He is currently an associate professor with the Institute for Network Sciences and Cyberspace, Tsinghua University. His research interests include network and system security, particularly in internet and cloud security, mobile security, and Big Data security.
\end{IEEEbiography}

\begin{IEEEbiography}[{\includegraphics[width=1in,height=1.25in,clip,keepaspectratio]{Quan_Z_Sheng.pdf}}]{Quan Z. Sheng} received the PhD
degree in computer science from the University of New South Wales and did his post-doc as a research scientist with CSIRO ICT Centre. He is a full professor and Head of School of Computing with Macquarie University. Before moving to Macquarie, he spent ten years at School of Computer Science, the University of Adelaide, serving in a number of senior leadership roles including acting Head and Deputy Head of School of Computer Science. His research interests include the Internet of Things, Big Data analytics, knowledge discovery, and Internet technologies. Professor Sheng is ranked by Microsoft Academic as one of the Most Impactful Authors in Services Computing (ranked Top 5 All Time). He is the recipient of the AMiner Most Influential Scholar Award on IoT (2007-2017), ARC Future Fellowship (2014), Chris Wallace Award for Outstanding Research Contribution (2012), and Microsoft Research Fellowship (2003).
\end{IEEEbiography}

\begin{IEEEbiography}[{\includegraphics[width=1in,height=1.25in,clip,keepaspectratio]{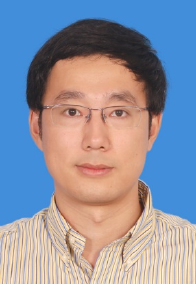}}]{Laizhong Cui}  is currently a Professor in the College of Computer Science and Software Engineering at Shenzhen University, China. He received the B.S. degree from Jilin University, Changchun, China, in 2007 and Ph.D. degree in computer science and technology from Tsinghua University, Beijing, China, in 2012. His research interests include Future Internet Architecture and Protocols, Edge Computing, Multimedia Systems and Applications, Blockchain, Internet of Things, Cloud Computing and Federated Learning. He led more than 10 scientific research projects, including National Key Research and Development Plan of China, National Natural Science Foundation of China, Guangdong Natural Science Foundation of China and Shenzhen Basic Research Plan. He has published more than 100 papers, including IEEE JSAC, IEEE TC, IEEE TPDS, IEEE TKDE, IEEE TMM, IEEE IoT Journal, IEEE TII, IEEE TVT, IEEE TNSM, ACM TOIT, IEEE Network, IEEE INFOCOM, ACM MM, IEEE ICNP, IEEE ICDCS, etc. He serves as an Associate Editor or a Member of Editorial Board for several international journals, including IEEE Internet of Things Journal, IEEE Transactions on Cloud Computing, IEEE Transactions on Network and Service Management, and International Journal of Machine Learning and Cybernetics. He is a Senior Member of the IEEE, and a Distinguished Member of the CCF.
\end{IEEEbiography}

\begin{IEEEbiography}[{\includegraphics[width=1in,height=1.25in,clip,keepaspectratio]{Jiangchuan_Liu.pdf}}]{Jiangchuan Liu} received the B.Eng.
degree (cum laude) in computer science from Tsinghua University, Beijing, China, in 1999, and the Ph.D. degree in computer science from The
Hong Kong University of Science and Technology in 2003. He was an Assistant Professor with The Chinese University of Hong Kong and a Research
Fellow at Microsoft Research Asia. He was also the EMC-Endowed Visiting Chair Professor of Tsinghua University from 2013 to 2016. He is currently a University Professor with the School of Computing Science, Simon Fraser University, Burnaby, BC, Canada. His research interests include multimedia systems and networks, cloud and edge computing, social networking, online gaming, and the Internet of Things/RFID/backscatter. Dr. Liu is a fellow of The Canadian Academy of Engineering and an NSERC E.W.R. Steacie Memorial Fellow. He was a co-recipient of the Inaugural Test of Time Paper Award of IEEE INFOCOM in 2015, the ACM SIGMM TOMCCAP Nicolas D. Georganas Best Paper Award in 2013, and the ACM Multimedia Best Paper Award in 2012. He was a Steering Committee Member of IEEE TRANSACTIONS ON MOBILE COMPUTING and the Steering Committee Chair of IEEE/ACM IWQoS from 2015 to 2017. He was the TPC Co-Chair of the IEEE INFOCOM in 2021. He has served on the editorial boards for IEEE/ACM TRANSACTIONS ON NETWORKING, IEEE TRANSACTIONS ON BIG DATA, IEEE TRANSACTIONS ON MULTIMEDIA, IEEE COMMUNICATIONS SURVEYS AND TUTORIALS, and IEEE INTERNET OF THINGS JOURNAL.
\end{IEEEbiography}

\clearpage

\appendices

\section{Notation List}
\label{appendix:notation}

 \begin{table}[!ht]
 \vspace{-1em}
		\begin{center}
			\begin{tabular}{|c|c|}
   \toprule[1pt] 
				  \text{Notation} & \text{Meaning} \\
    \midrule $T$ & the total global iterations\\
    \hline $T_0$ & the total global iterations for parameter estimation\\
     \hline $N$ & the total number of clients\\
      \hline $K$ &  \makecell*[c]{the selected number of clients \\in each global iteration}\\
       \hline
    $\mathcal{N}$ & the total client set\\
      \hline
    $\mathcal{S}_t$ & the selected client set in the $t$-th global iteration\\
   
				\hline $(\epsilon_n, \delta_n)$& the privacy budget of client $n$\\
				\hline
                $p_n$& the selection probability of client $n$\\
                \hline
                $T_n$& the total selection times of client $n$\\
                \hline
                $\Gamma_n$& the non-IID degree of client $n$\\
				\hline
    $C_n$& the selected times of client $n$\\
    \hline
     $\hat{\mathbf{w}}^t$& the global model in the $t$-th global iteration\\
     \hline
      $\mathbf{w}^t_n(\hat{\mathbf{w}}^t_n)$& \makecell*[c]{the local model (disturbed by  DP noises) \\ of client $n$ in the $t$-th global iteration}\\
				\hline
    $\mathbf{g}^t_n (\hat{\mathbf{g}}^t_n)$& \makecell*[c]{the local gradients (disturbed by  DP noises) \\of client $n$ in the $t$-th global iteration}\\
    \hline
    $\eta_t$& the learning rate in the $t$-th global iteration\\
    \hline
    $\mathbf{Z}^t_n$ & \makecell*[c]{the DP noises generated by client $n$ to disturb \\gradients in the $t$-th global iteration}\\
    
    \hline
     $\Xi_G(\Xi_L)$& the $L2(L1)$ magnitude
bound of gradients\\
  
    \bottomrule[1pt]
			\end{tabular}
		\end{center}
		\caption{Notation List} 
		\label{table+notation}
\end{table}
 
\section{Proof of Theorem \ref{the:convergence}}	

\label{proof_theorem}

\begin{proof}

The convergence of FL under convex loss and random client selection has been derived in \cite{DBLP:conf/iclr/LiHYWZ20}. In our study, there are three differences attributed to DP noises and biased client selection. More detailed proof is presented as follows.

We define the expectation of $\hat{\mathbf{g}}^t_n$ as $\bar{\mathbf{g}}^t_n=\mathbb{E}[\hat{\mathbf{g}}^t_n]=\mathbb{E}[\mathbf{g}^t_n]+\mathbb{E}[\mathbf{Z}^t_n]=\eta_t \nabla F_n(\hat{\mathbf{w}}^{t-1})$. According to \cite{DBLP:conf/iclr/LiHYWZ20}, we have

\begin{align}
		&\mathbb{E}\| \hat{\mathbf{w}}^{t}-\mathbf{w}^*\|^2\le(1-\mu\eta_t)\mathbb{E}\| \hat{\mathbf{w}}^{t}-\mathbf{w}^*\|^2\notag\\
  +&\underbrace{\mathbb{E}\| \frac{1}{K}\sum_{n\in\mathcal{S}_t}(\bar{\mathbf{g}}^t_n-\hat{\mathbf{g}}^t_n)\|^2}_{A_1}
  +\underbrace{\frac{2L\eta_t^2}{K}\mathbb{E}\sum_{n\in\mathcal{S}_t}[F_n(\mathbf{w}^*)-F_n^*]}_{A_2}\notag\\+&\underbrace{\frac{2\eta_t}{K}(L\eta_t-1)\mathbb{E}\sum_{n\in\mathcal{S}_t}[F_n(\hat{\mathbf{w}}^{t-1}) - F_n(\mathbf{w}^*)]}_{A_3}.
	\end{align}

	For $A_1$, we have
 \begin{align}
		&\mathbb{E}\| \frac{1}{K}\sum_{n\in\mathcal{S}_t}(\bar{\mathbf{g}}^t_n-\hat{\mathbf{g}}^t_n)\|^2
		=\frac{1}{K}\sum_{n\in \mathcal{N}}p_n\mathbb{E}\| \bar{\mathbf{g}}^t_n-\mathbf{g}^t_n-\mathbf{Z}^t_n\|^2\notag\\
		\le&\frac{2}{K}\sum_{n\in \mathcal{N}}p_n(\mathbb{E}\| \bar{\mathbf{g}}^t_n-\mathbf{g}^t_n\|^2+\mathbb{E}\|\mathbf{Z}^t_n\|^2)\notag\\
		\le&\frac{2\eta_t^2\sigma^2}{K}+\frac{2\eta_t^2\Lambda}{K^2T}\sum_{n\in \mathcal{N}}T_n^{z+1}\Phi_n,
		\label{eq:a4}
	\end{align}which follows  $p_n =\frac{T_n}{KT}$, $\| \mathbf{a}+\mathbf{b}\|^2\le 2\| \mathbf{a}\|^2+2\| \mathbf{b}\|^2$, Assumption \ref{ass:3},  Lemma \ref{lem:Gaussian_variance} and Lemma \ref{lem:Laplace variance}. 
 
 For $A_2$, with $p_n = \frac{T_n}{KT}$, we have
	\begin{align}
		\frac{2L\eta_t^2}{K}\mathbb{E}\sum_{n\in\mathcal{S}_t}[F_n(\mathbf{w}^*)-F_n^*] \le \frac{2L\eta_t^2}{KT}\sum_{n\in\mathcal{N}}T_n\Gamma_n.
		\label{eq:C2}
	\end{align}

	For $A_3$, with $p_n = \frac{T_n}{KT}$, we have
	\begin{align}
		&\frac{2\eta_t}{K}(L\eta_t-1)\mathbb{E}\sum_{n\in\mathcal{S}_t}[F_n(\hat{\mathbf{w}}^{t-1}) - F_n(\mathbf{w}^*)]\notag\\
		\overset{(a)}{\le}& \frac{3\eta_t}{2}\sum_{n\in\mathcal{N}}\frac{T_n}{KT}[F_n(\mathbf{w}^*)-F_n^*]\notag\\&-\frac{3\eta_t\rho(\hat{\mathbf{w}}^{t-1})}{2}[F(\hat{\mathbf{w}}^{t-1})-\frac{1}{N}\sum_{n\in\mathcal{N}}F^*_n]\notag\\
		\overset{(b)}{\le}&\frac{3\eta_t}{2}\sum_{n\in\mathcal{N}}(\frac{T_n}{KT}-\frac{\rho_{min}}{N})\Gamma_n,
		\label{eq:C1}
	\end{align}where $(a)$ follows $\eta_t \le \frac{1}{4L}$ and Definition \ref{def:rho}; $(b)$ follows Definition \ref{def:rho_min}, $F(\hat{\mathbf{w}}^{t-1})\ge F^* = \frac{1}{N}\sum_{n\in\mathcal{N}}F_n(\mathbf{w}^*)$ and $\Gamma_n = F_n(\mathbf{w}^*)-F_n^*$.

	According to Eqs.~(\ref{eq:a4}), (\ref{eq:C2}) and (\ref{eq:C1}), and let $\Delta_t=	\mathbb{E}\| \hat{\mathbf{w}}^{t}-\mathbf{w}^*\|^2$, $\Upsilon_A=\frac{2\sigma^2}{K}+\frac{2\Lambda}{K^2T}\sum_{n\in\mathcal{N}}T_n^{z+1}\Phi_n+\frac{2L}{KT}\sum_{n\in\mathcal{N}}T_n\Gamma_n$ and $\Upsilon_B=\frac{3}{2}\sum_{n\in\mathcal{N}}(\frac{T_n}{KT}-\frac{\rho_{min}}{N})\Gamma_n$, we have 
	\begin{align}
		\Delta_{t}\le(1-\mu\eta_t)\Delta_{t-1}+\eta_t^2\Upsilon_A+\eta_t\Upsilon_B.
	\end{align}

	By setting $\Delta_t\le\frac{\psi}{t+\gamma}$, $\eta_t=\frac{\beta}{t+\gamma}$ where $\beta>\frac{1}{\mu }$ and $\gamma>0$, we have
	\begin{align}
		\psi=\max\{\gamma\| \hat{\mathbf{w}}^0-\mathbf{w}^*\|^2,\frac{\beta^2\Upsilon_A+\Upsilon_B\beta(\gamma+t)}{\beta\mu  -1}\}.
	\end{align}
	
	Then, by the $L$-smooth of $F$, we have
	\begin{align}
		\mathbb{E}[F(\hat{\mathbf{w}}^t)]-F^*\le\frac{L}{2}\Delta_t\le\frac{L}{2}\frac{\psi}{\gamma+t}.
	\end{align}
	
	With $\beta=\frac{2}{\mu}$ and $\gamma=\frac{8L}{\mu}$, we have

	\begin{align}
		&\mathbb{E}[F(\hat{\mathbf{w}}^T)]-F^*\notag\\
		\le&\frac{L}{2(\gamma+T)}(\gamma\| \hat{\mathbf{w}}^0-\mathbf{w}^*\|^2+\frac{\beta^2\Upsilon_A+\Upsilon_B\beta(\gamma+T)}{\beta\mu  -1})\notag\\
		=&\frac{1}{\gamma+T}\frac{L\gamma}{2}\| \hat{\mathbf{w}}^0-\mathbf{w}^*\|^2+\frac{1}{\gamma+T}\frac{4L\sigma^2}{\mu^2K}\notag\\
		+&\frac{1}{(\gamma+T)T}\frac{4L^2}{K\mu^2}\sum_{n\in\mathcal{N}}T_n\Gamma_n+\frac{3L}{2\mu}\sum_{n\in\mathcal{N}}(\frac{T_n}{KT}-\frac{\rho_{min}}{N})\Gamma_n\notag\\+&\frac{1}{(\gamma+T)T}\frac{4L\Lambda}{K^2\mu^2}\sum_{n\in\mathcal{N}}T_n^{z+1}\Phi_n.\end{align}\end{proof}
	

\section{Proof of Theorem \ref{the:convex_function}}	
\label{appendix: proof_of_convex_function}
\begin{proof}
    It is easy to obtain the Hessian Matrix of the objective function $\mathcal{J}(T_1, \cdots, T_N) = \Omega_A\sum_{n\in\mathcal{N}}T_n^{z+1}\Phi_n+\Omega_B\sum_{n\in\mathcal{N}}T_n\Gamma_n$  by performing second-order partial derivatives with respect to $T_n, \forall n\in\mathcal{N}$. Obviously, the matrix is a diagonal matrix because $\frac{\partial^2 \mathcal{J}}{\partial T_n\partial T_{n'}} = 0$ if $n\neq n'$. In the matrix,  diagonal elements, \emph{i.e.}, $\frac{\partial^2 \mathcal{J}}{\partial T_n^2}=(z+1)z\Omega_A\Phi_nT_n^{z-1}$ ($z=1$ for Gaussian mechanism and $z=2$ for Laplace mechanism), are greater than or equal to 0. It means that all its eigenvalues are greater than or equal to 0. Therefore, it is easy to prove that the matrix is a semi positive-definite matrix, and hence  $\mathcal{J}(\cdot)$ is a convex function with respect to $T_n$. 
\end{proof}

\section{Proof of Proposition \ref{Prop:ApproxSolu}}	
\label{appendix: proof_of_ApproxSolu}


\begin{proof}
    With $\Gamma_n = 0, \forall n\in\mathcal{N}$, the objective function in $\mathbb{P}_1$ can be simplified as $\mathcal{J}'(T_1, \cdots, T_N) = \sum_{n\in\mathcal{N}}T_n^{z+1}\Phi_n $. If we relax the constraint by letting $T_n$ be a non-negative real number, it is easy to prove that $\mathcal{J}'(T_1, \cdots, T_N)$ is also a convex objective function with respect to $T_n$ with both the  Gaussian and Laplace mechanisms. Therefore, we can obtain the optimal analytical solution to minimize $\mathcal{J}'(\cdot)$  by  using Lagrange Multiplier. By introducing a lagrange multiplier $\lambda$, the equality constraint, $\sum_{n\in\mathcal{N}}T_n = KT$, can be transformed into a part of the new objective function, represented as $\mathcal{J}''(T_1, \cdots, T_N, \lambda) = \sum_{n\in\mathcal{N}}T_n^{z+1}\Phi_n + \lambda (\sum_{n\in\mathcal{N}}T_n - KT)$.

Then, we can get the partial derivative of $\mathcal{J}''(\cdot)$ with respect to $T_n, \forall n\in\mathcal{N}$, which is represented as $\frac{\partial \mathcal{J}''}{\partial T_n} = (z+1)T_n^z\Phi_n + \lambda$, and with respect to $\lambda$, which is represented as $\frac{\partial \mathcal{J}''}{\partial \lambda} = \sum_{n\in\mathcal{N}}T_n - KT$. Finally, we can obtain the optimal solution, $T_{n,0}^*, \forall n\in\mathcal{N}$, to minimize $\mathcal{J}''(\cdot)$, which satisfies $\frac{\partial \mathcal{J}''}{\partial T_{n,0}^*} = 0, \forall n\in\mathcal{N}$, and $\frac{\partial \mathcal{J}''}{\partial \lambda} = 0$. 

With $\frac{\partial \mathcal{J}''}{\partial T_{n,0}^*} = 0, \forall n\in\mathcal{N}$, for any client $i$ and $j$ ($i, j\in\mathcal{N}$), we have $T_{i,0}^* = (\frac{\Phi_j}{\Phi_i})^{(\frac{1}{z})}T_{j,0}^*$. Applying the above equations into equation $\sum_{n\in\mathcal{N}} T_{n,0}^* - KT = 0$, we can obtain the optimal solution of each client $n$ by solving $\sum_{n'\in\mathcal{N}}(\frac{\Phi_n}{\Phi_{n'}})^{\frac{1}{z}}T_{n,0}^*-KT = 0$, that is
$T_{n,0}^* =\frac{KT}{\sum_{n'\in\mathcal{N}}(\frac{\Phi_n}{\Phi_{n'}})^{\frac{1}{z}}}$.\end{proof}

\section{Parameter Estimation and Decision}

\label{appendix:problem-related}
\begin{figure}[ht]
  
 \includegraphics[width =\linewidth]{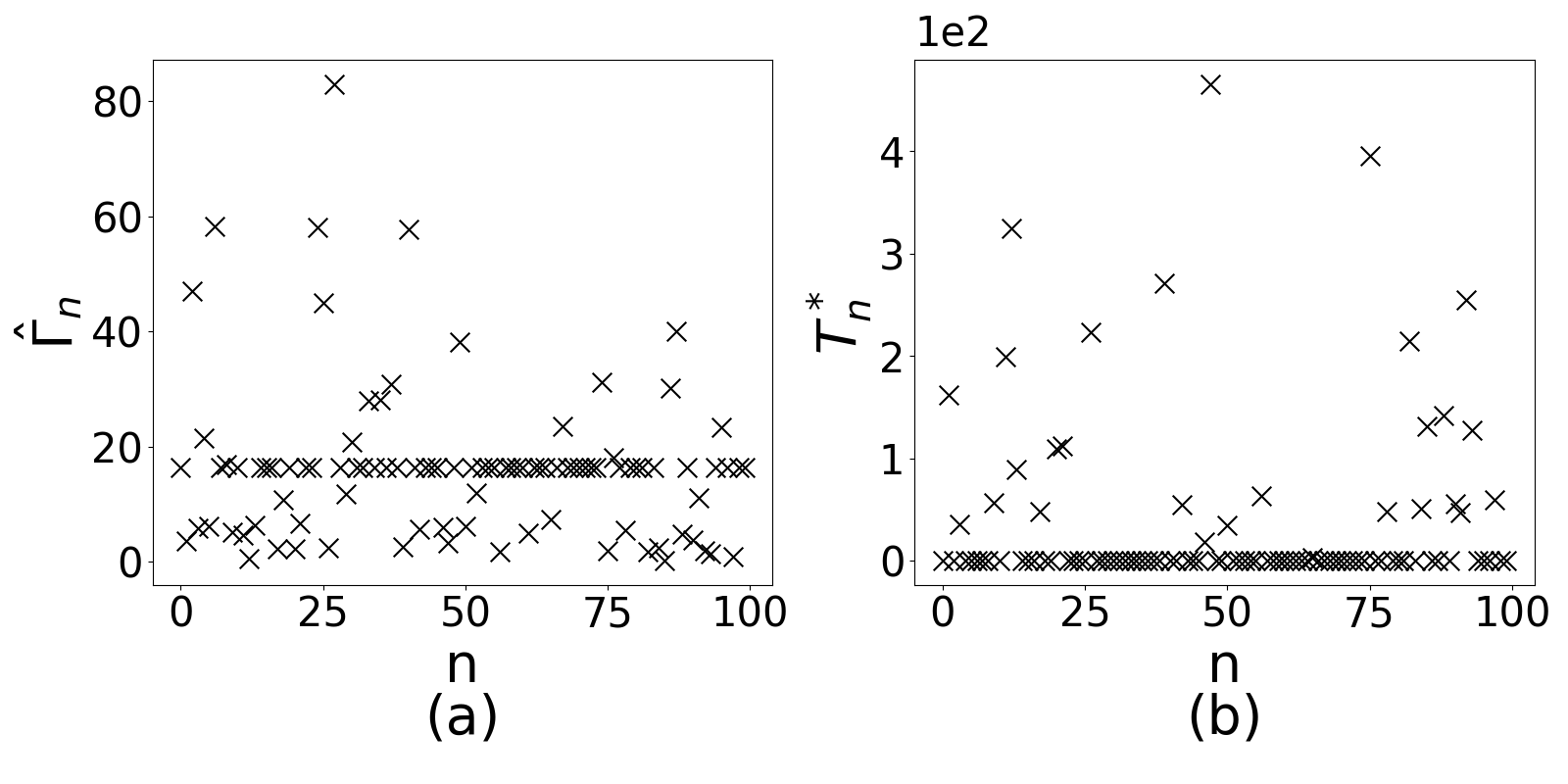}

    \caption{The estimation of problem-related parameter $\Gamma_n$ and the optimal client selection decision $T_n^*$ in Lending Club with Gaussian Mechanism.}
\label{fig:lending+Gaussian+all+est}

\end{figure}

\begin{figure}[!ht]
 \includegraphics[width =\linewidth]{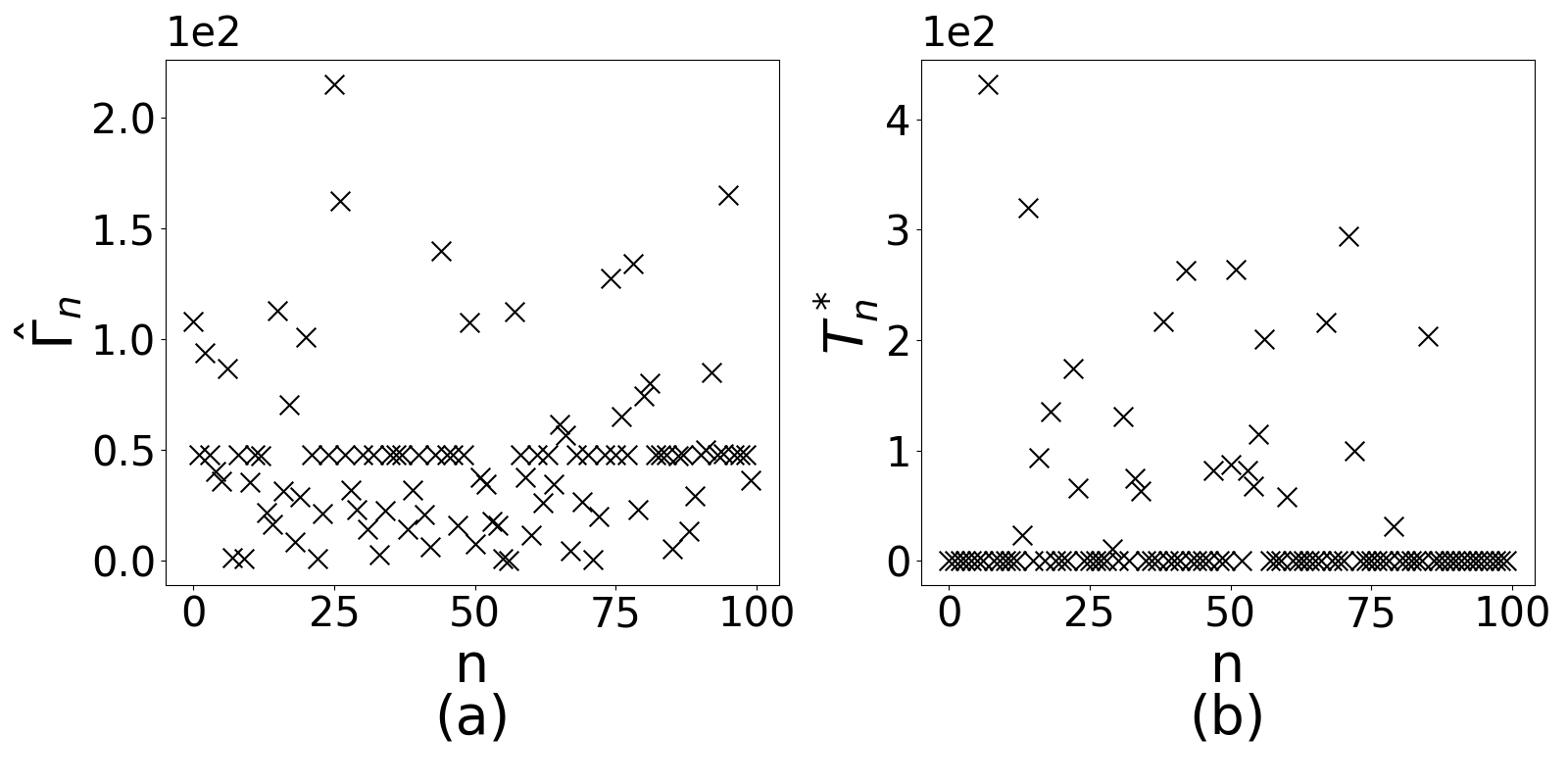}

    \caption{The estimation of problem-related parameter $\Gamma_n$ and the optimal client selection decision $T_n^*$ in Lending Club with Laplace Mechanism.}
\label{fig:lending+Laplace+all+est}
 
\end{figure}

\begin{figure}

 \includegraphics[width =\linewidth]{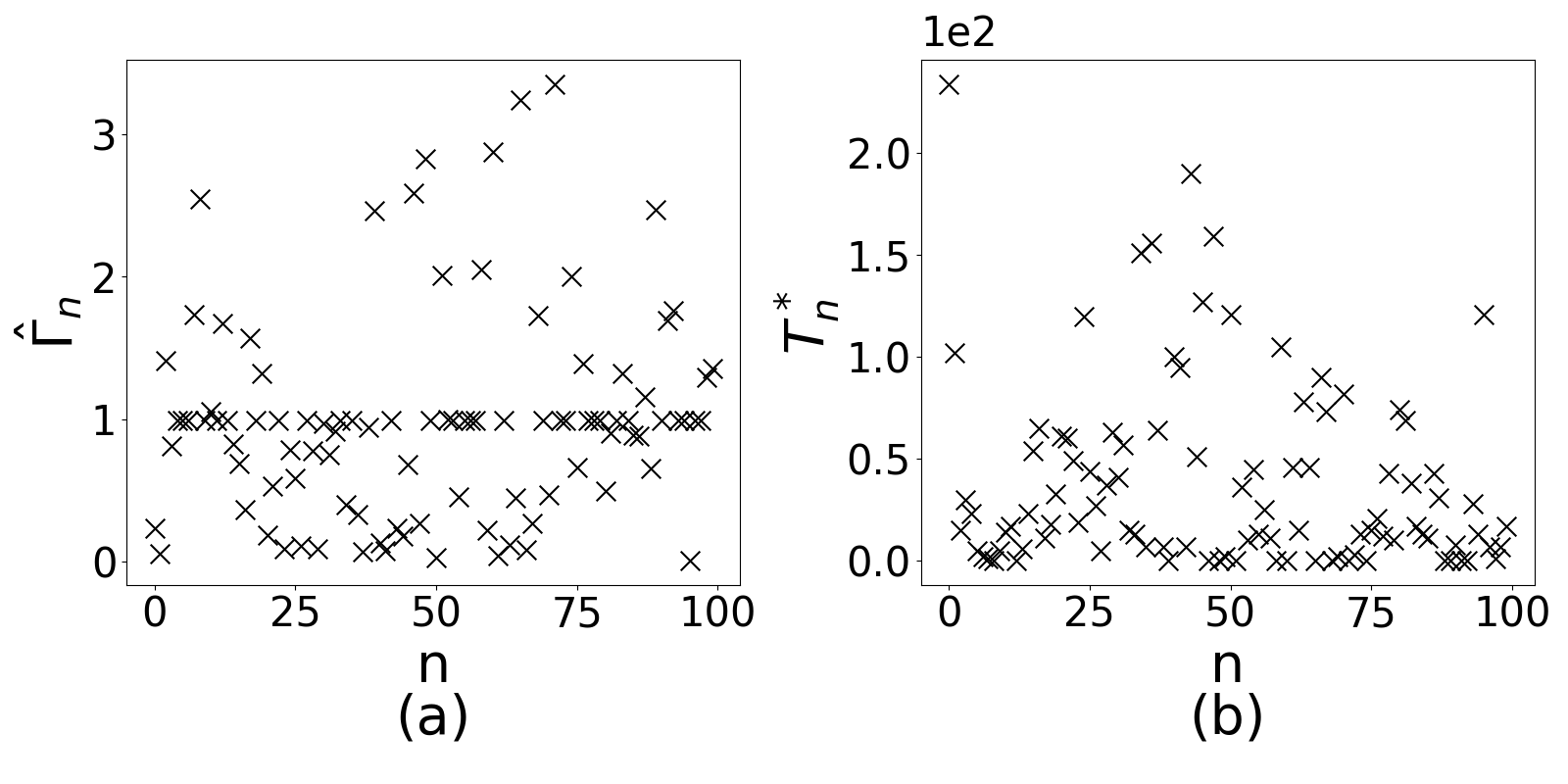}

    \caption{The estimation of problem-related parameter $\Gamma_n$ and the optimal client selection decision $T_n^*$ in MNIST with Laplace Mechanism.}
\label{fig:mnist+Laplace+all+est}
\end{figure}

\begin{figure}
 \includegraphics[width =\linewidth]{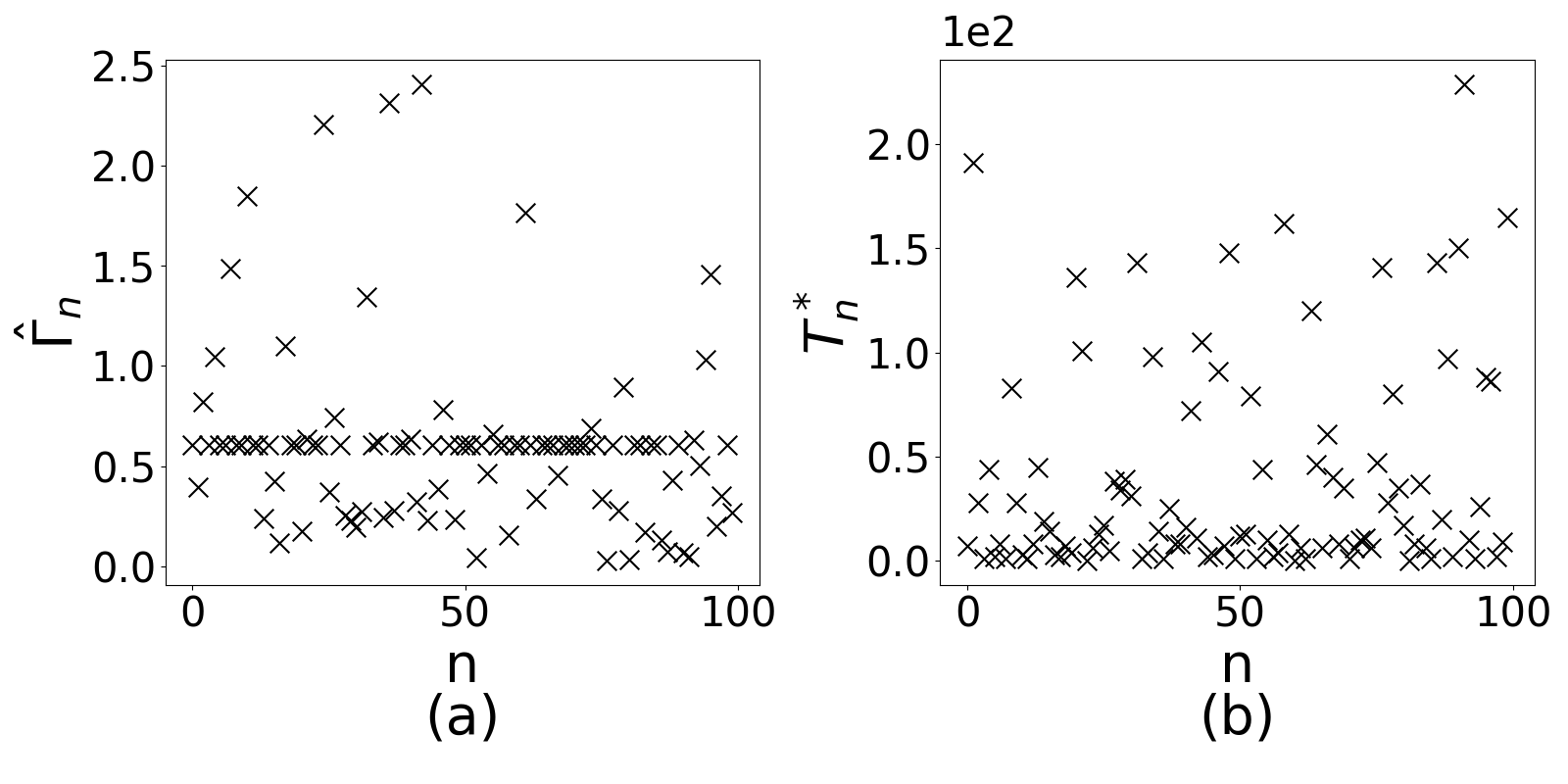}
 
    \caption{The estimation of problem-related parameter $\Gamma_n$ and the optimal client selection decision $T_n^*$ in Fashion-MNIST with Gaussian Mechanism.}
\label{fig:fashionmnist+Gaussian+all+est}
\end{figure}

\begin{figure}
 \includegraphics[width =\linewidth]{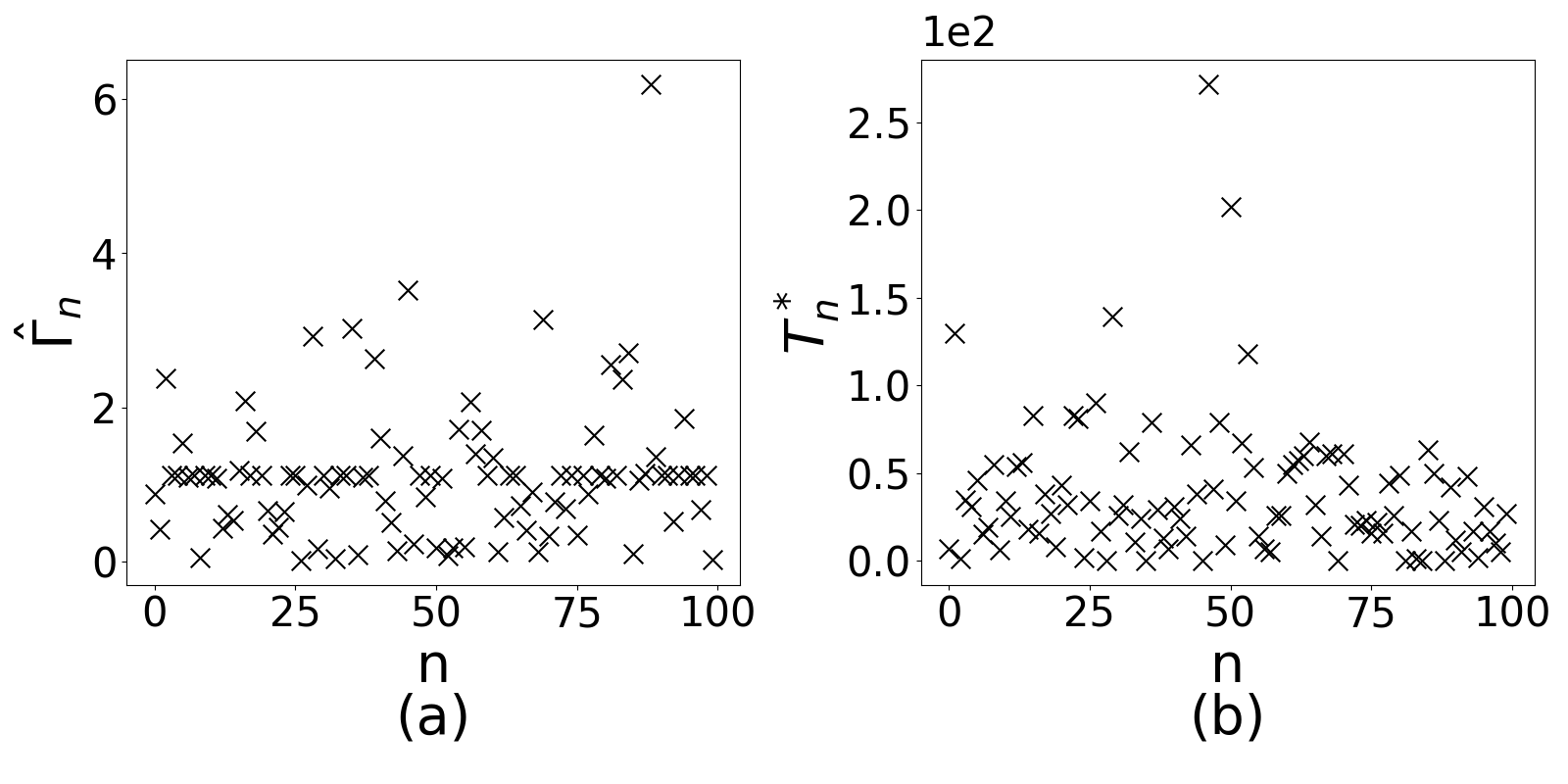}
  
    \caption{The estimation of problem-related parameter $\Gamma_n$ and the optimal client selection decision $T_n^*$ in Fashion-MNIST with Laplace Mechanism.}
\label{fig:fashionmnist+Laplace+all+est}
\end{figure}

\begin{figure}
 \includegraphics[width =\linewidth]{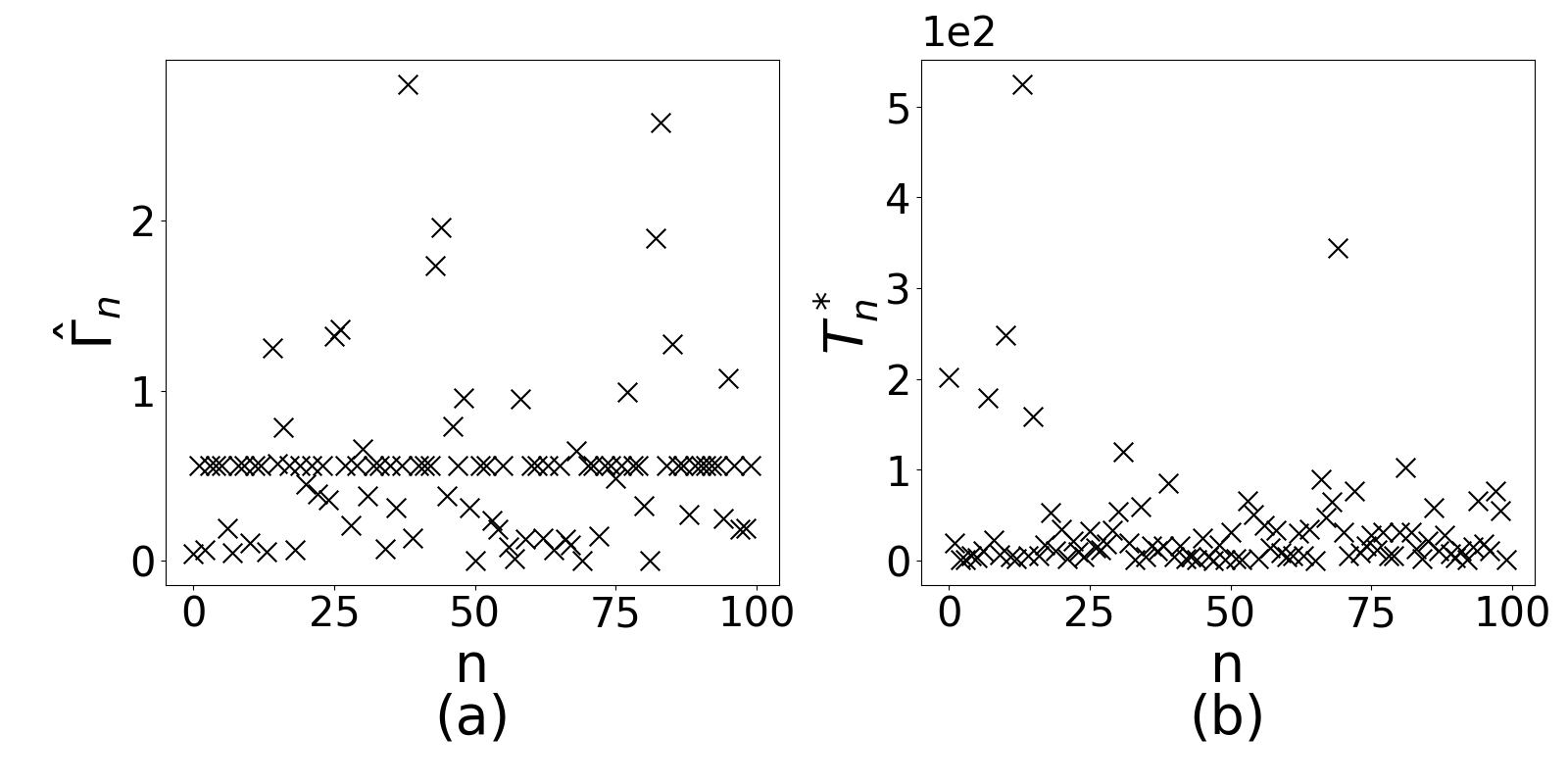}
  
    \caption{The estimation of problem-related parameter $\Gamma_n$ and the optimal client selection decision $T_n^*$ in FEMNIST with Gaussian Mechanism.}
\label{fig:femnist+Gaussian+all+est}
\end{figure}

\begin{figure}
 \includegraphics[width =\linewidth]{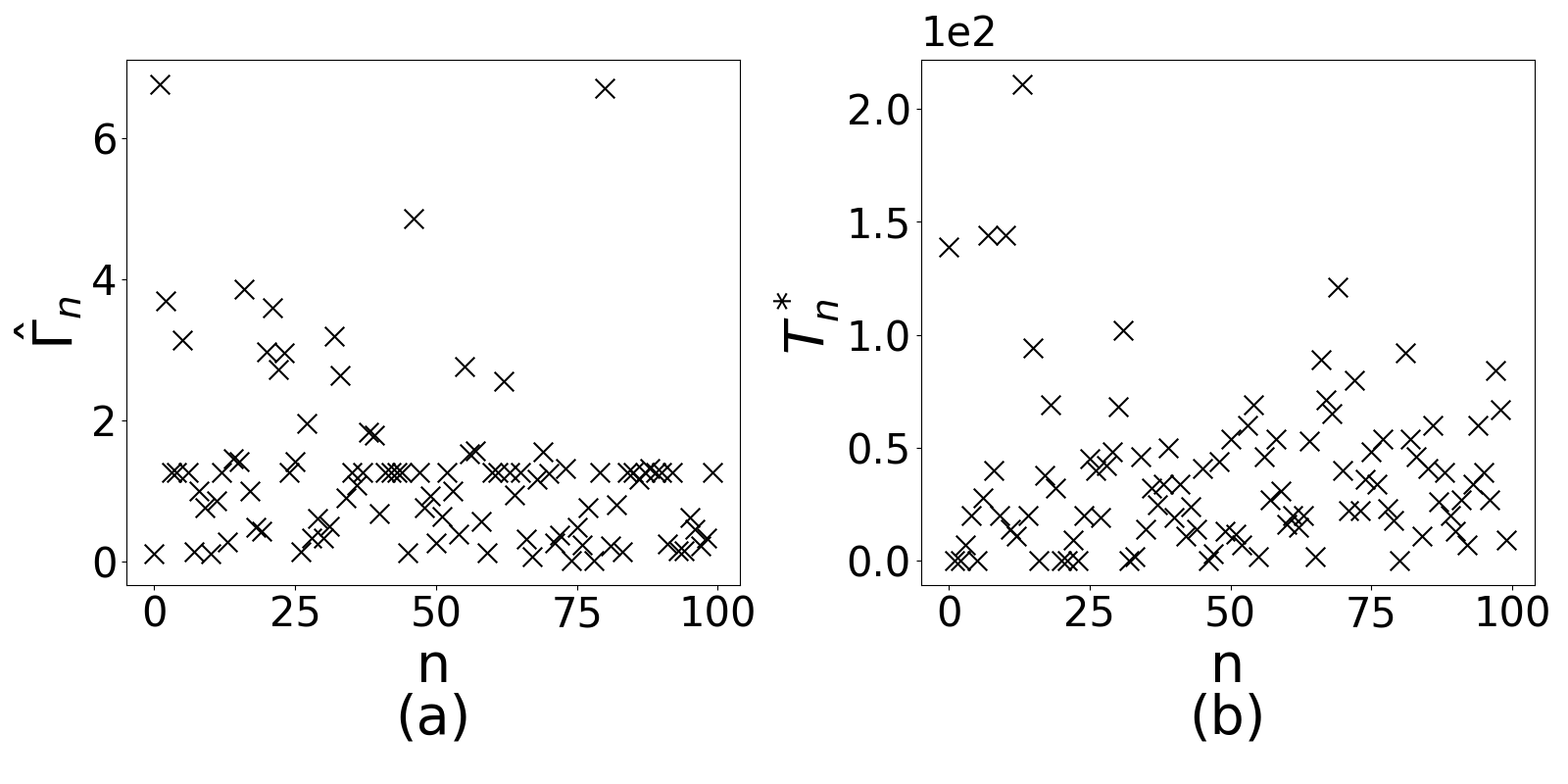}
  
    \caption{The estimation of problem-related parameter $\Gamma_n$ and the optimal client selection decision $T_n^*$ in FEMNIST with Laplace Mechanism.}
\label{fig:femnist+Laplace+all+est}
\end{figure}

\begin{figure}[!ht]
 \includegraphics[width =\linewidth]{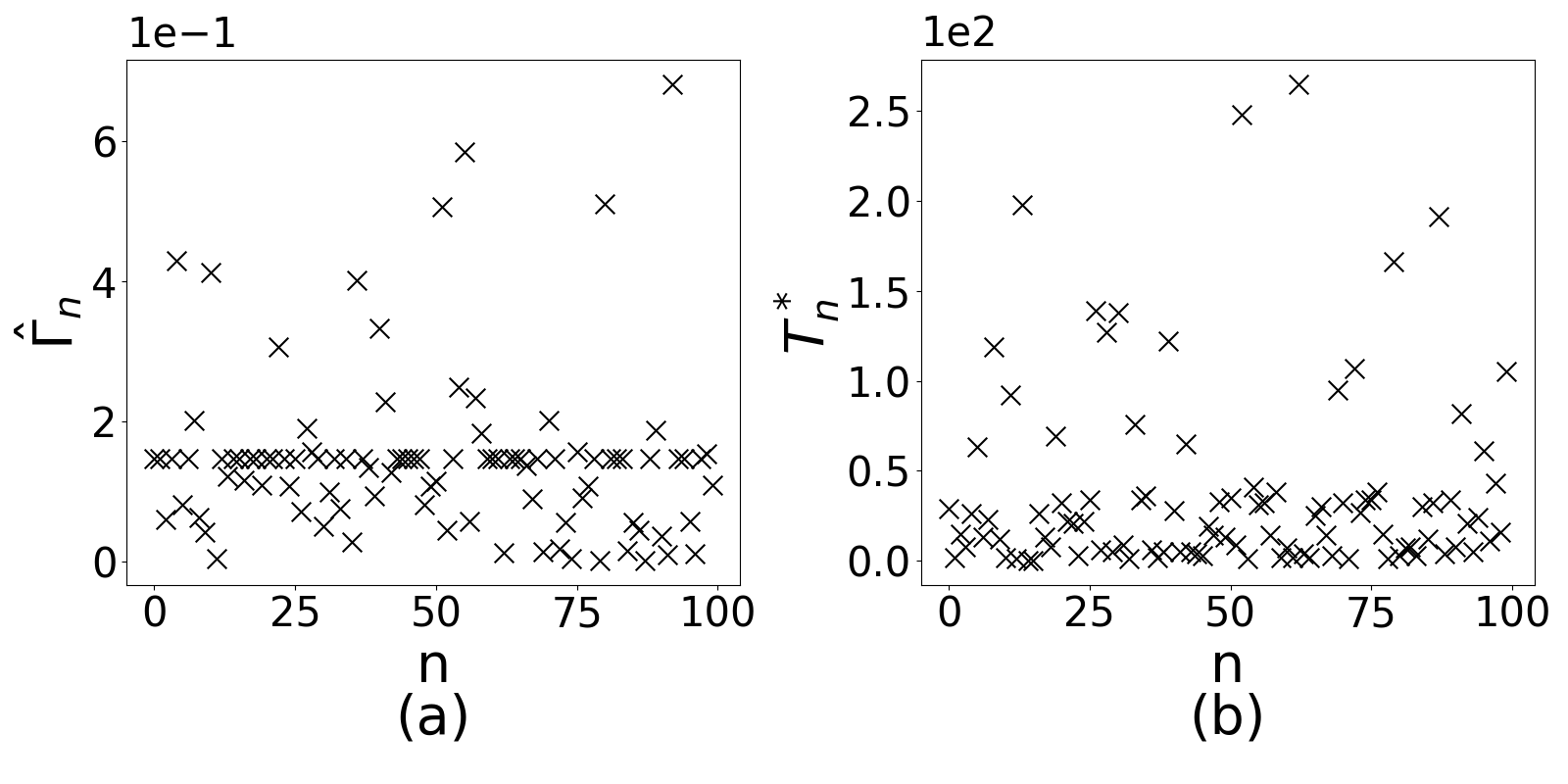}

    \caption{The estimation of problem-related parameter $\Gamma_n$ and the optimal client selection decision $T_n^*$ in CIFAR-10 with Gaussian Mechanism.}
\label{fig:cifar10+Gaussian+all+est}
\end{figure}

\begin{figure}[!ht]

 \includegraphics[width =\linewidth]{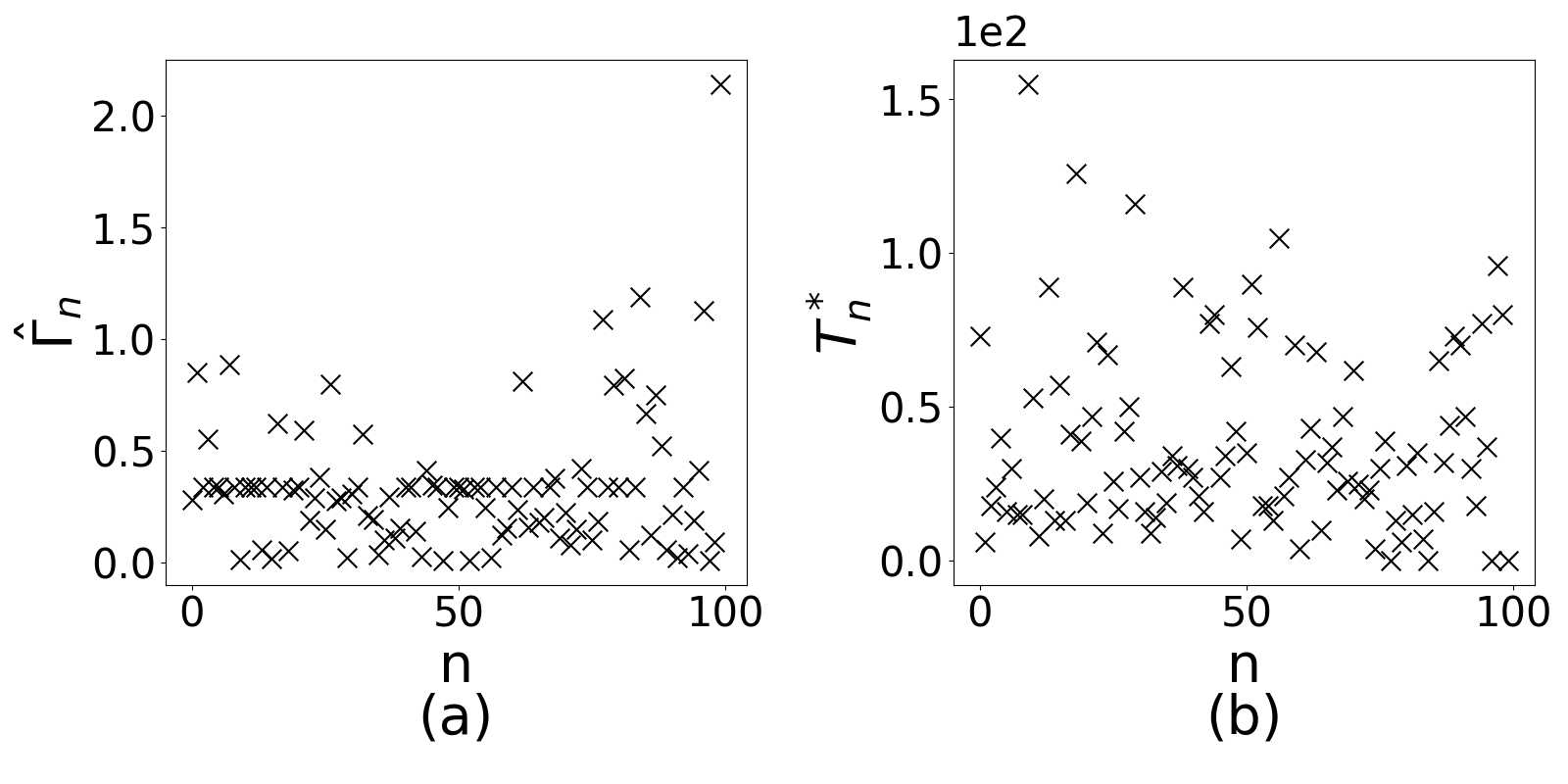}

    \caption{The estimation of problem-related parameter $\Gamma_n$ and the optimal client selection decision $T_n^*$ in CIFAR-10 with Laplace Mechanism.}
\label{fig:cifar10+Laplace+all+est}
\end{figure}

We visualize the estimation of problem-related parameter $\Gamma_n$ and the algorithmic decision $T_n^*$, similar to what we have presented in Fig.~\ref{fig:mnist+Gaussian+all+est}, for other cases with different datasets and DP mechanisms in Figs.~\ref{fig:lending+Gaussian+all+est}-\ref{fig:cifar10+Laplace+all+est}. With the estimation of $\Gamma_n$ and $\rho_{min}$, we can solve  problem $\mathbb{P}_2$ to obtain the estimation  of $\gamma$, $L$ and $\mu$ for each experiment case, which are shown in Table.~\ref{table+estimation}.





\clearpage

\newpage

\begin{table}[!ht]
		\begin{center}
			\begin{tabular}{c|c|cccc}
				\toprule[1pt]  \text{Dataset} & \text{DP} & $\hat{\rho}_{min}$& $\hat{\gamma}$  &$\hat{L}$  & $\hat{\mu}$  \\
				\midrule
				\multirow{2}{*}{\text{Lending}}& \text{GM }& 0.937&
				 0.277&0.997 &0.239 \\
				 \multirow{2}{*}{}& \text{LM }& 1.040&
				 1.046&3.472 &1.063 \\
				\hline 
				\multirow{2}{*}{\text{MNIST}}& \text{GM }& 1.063&
			1.960 &5.514 &4.986 \\
				\multirow{2}{*}{}& \text{LM }& 1.039&
				53.565 &106.463 &55.018 \\
				\hline
				\multirow{2}{*}{\text{F-MNIST}}& \text{GM }&1.050&
			12.117 &18.062 &14.198 \\
				\multirow{2}{*}{}& \text{LM }&0.962&
			144.688 &406.756 &200.645 \\
   \hline 
				\multirow{2}{*}{\text{FEMNIST}}& \text{GM }&1.037&
			5.971 &11.815 &10.515 \\
				\multirow{2}{*}{}& \text{LM }&1.053&
			118.286 &310.663 &153.461 \\
				\hline 
				\multirow{2}{*}{\text{CIFAR-10}}& \text{GM }&1.097&
				1.958 &5.859 &5.678 \\
				\multirow{2}{*}{}& \text{LM }&1.109&
			66.112 &125.825 &63.722 \\
				\bottomrule[1pt]
			\end{tabular}
		\end{center}
		\caption{The estimation of problem-related parameters with different datasets and different DP mechanisms. (Lending = Lending Club, F-MNIST=Fashion-MNIST, GM = Gaussian Mechanism, LM = Laplace Mechanism)} 
	\label{table+estimation}
	\end{table}

\balance

\end{document}